\documentclass[10pt]{article}

\PassOptionsToPackage{hypertexnames=false}{hyperref}

\usepackage[accepted]{tmlr}

\usepackage[utf8]{inputenc}             
\usepackage[T1]{fontenc}                
\usepackage{booktabs}                   
\usepackage{nicefrac}                   
\usepackage{microtype}                  
\usepackage{url}                        
\usepackage{xcolor}
\usepackage{xkcdcolors}
\usepackage{bookmark}                   
\usepackage{float}

\usepackage{algorithm}
\usepackage{algorithmic}

\usepackage{multirow}
\usepackage{wrapfig}

\usepackage{pgfplots}

\pgfplotsset{compat=1.18}

\usepackage{caption}
\makeatletter
\newcommand{\algcaption}[1]{\@namedef{caption@type@warning}{}\captionof{algorithm}{#1}}
\makeatother

\usepackage{enumitem}
\setlist[itemize]{noitemsep,left=7pt,nosep}

\usepackage[hypertexnames=false]{hyperref}
\hypersetup{
  pdftitle={Generative Modeling with Bayesian Sample Inference},
  pdfauthor={Marten Lienen, Marcel Kollovieh, Stephan Günnemann},
  pdfsubject={Published in Transactions on Machine Learning Research (08/2026)},
  colorlinks=true,
  linkcolor=xkcdCrimson,
  urlcolor=xkcdBluish,
  citecolor=xkcdLightNavy,
}

\usepackage{amsmath}
\usepackage{amssymb}
\usepackage{amsfonts} 
\usepackage{amsthm}
\usepackage{mathtools}
\usepackage{thmtools,thm-restate}

\usepackage[mode=match,uncertainty-mode=separate]{siunitx}
\DeclareSIUnit{\million}{M}
\DeclareSIUnit{\thousand}{k}

\usepackage[shortcuts=abbr]{glossaries-extra}
\glsdisablehyper{}

\usepackage[status=final,layout=inline,author=TODO]{fixme}
\fxusetheme{color}

\usepackage{amsmath,amsfonts,bm}

\def\eqref#1{equation~\ref{#1}}

\def\1{\bm{1}}

\def\vzero{{\bm{0}}}

\def\vmu{{\bm{\mu}}}
\def\vnu{{\bm{\nu}}}

\def\vtheta{{\bm{\theta}}}

\def\vx{{\bm{x}}}
\def\vy{{\bm{y}}}
\def\vz{{\bm{z}}}

\def\evx{{x}}

\def\mI{{\bm{I}}}

\DeclareMathAlphabet{\mathsfit}{\encodingdefault}{\sfdefault}{m}{sl}
\SetMathAlphabet{\mathsfit}{bold}{\encodingdefault}{\sfdefault}{bx}{n}

\def\sN{{\mathbb{N}}}

\def\sS{{\mathbb{S}}}

\newcommand{\R}{\mathbb{R}}

\makeatletter
\patchcmd\WF@putfigmaybe{\lower\intextsep}{}{}{\fail}%
\AddToHook{env/wraptable/begin}{\setlength{\intextsep}{0pt}}
\makeatother

\usepackage[capitalise]{cleveref}

\usepackage{crossreftools}
\let\ORGhypersetup\hypersetup
\protected\def\hypersetup{\ORGhypersetup}
\newcommand\extrafootertext[1]{%
  \bgroup%
  \renewcommand\thefootnote{\fnsymbol{footnote}}%
  \renewcommand\thempfootnote{\fnsymbol{mpfootnote}}%
  \footnotetext[0]{#1}%
  \egroup%
}

\theoremstyle{plain}
\newtheorem{theorem}{Theorem}[section]

\newtheorem{lemma}[theorem]{Lemma}

\theoremstyle{definition}

\theoremstyle{remark}

\usepackage{accents}

\renewcommand\underbar[1]{\underaccent{\bar}{#1}}

\DeclareMathSymbol{\shortminus}{\mathbin}{AMSa}{"39}

\newcommand{\vpsi}{\bm{\psi}}
\newcommand{\vepsilon}{\bm{\varepsilon}}
\newcommand{\pr}{\mathrm{p}}

\newcommand{\qr}{\mathrm{q}}
\newcommand{\N}{\mathcal{N}}
\newcommand{\LogUniform}{\operatorname{Log-Uniform}}
\newcommand{\LUcdf}{F}
\newcommand{\Np}{\mathcal{N}_{\mathrm{P}}}
\newcommand{\Rplus}{\mathbb{R}_{+}}

\newcommand{\T}{^{\mathsf{T}}}

\newcommand{\inv}{^{\shortminus{}\hspace{-0.075em}1}}

\newcommand{\rsqrt}{^{\shortminus{}\hspace{-0.075em}\nicefrac{1}{2}}}
\newcommand{\bigprsqrt}{^{\negthickspace\shortminus{}\hspace{-0.075em}\nicefrac{1}{2}}}
\newcommand{\powsqrt}{^{\nicefrac{1}{2}}}
\newcommand{\kl}{D_{\mathrm{KL}}}

\DeclareMathOperator*{\E}{\mathbb{E}}
\DeclareMathOperator{\var}{\mathrm{Var}}

\newcommand{\ndim}{n}
\newcommand{\nrounds}{k}

\newcommand{\nbuckets}{r}
\newcommand{\vxhat}{\hat{\vx}}
\newcommand{\model}{f_{\vtheta}}

\newcommand{\alphaR}{\alpha_{\mathrm{R}}}
\newcommand{\alphaM}{\alpha_{\mathrm{M}}}
\newcommand{\lambdaM}{\lambda_{\mathrm{M}}}
\newcommand{\lrecon}{\mathcal{L}_{\mathrm{R}}}
\newcommand{\lmeasurek}{\mathcal{L}_{\mathrm{M}}^{k}}
\newcommand{\lmeasurekprime}{\mathcal{L}_{\mathrm{M}}^{k'}}
\newcommand{\lmeasureinf}{\mathcal{L}_{\mathrm{M}}^{\infty}}
\newcommand{\preskip}{c_{\mathrm{skip}}}
\newcommand{\preout}{c_{\mathrm{out}}}
\newcommand{\prein}{c_{\mathrm{in}}}

\newcommand{\defabbrcmd}[1]{
  \expandafter\def\csname#1\endcsname{{\gls{#1}}}
}
\newcommand{\defabbrcmds}[1]{
  \expandafter\def\csname#1\endcsname{{\gls{#1}}}
  \expandafter\def\csname#1s\endcsname{{\glspl{#1}}}
}

\newabbreviation{elbo}{ELBO}{evidence lower bound}
\defabbrcmds{elbo}
\newabbreviation{bpd}{BPD}{bits per dimension}
\defabbrcmd{bpd}
\newabbreviation{mc}{MC}{Monte Carlo}
\defabbrcmd{mc}
\newabbreviation{bsi}{BSI}{Bayesian Sample Inference}
\defabbrcmd{bsi}
\newabbreviation{bfn}{BFN}{Bayesian Flow Network}
\defabbrcmds{bfn}
\newabbreviation{pmm}{PMM}{Posterior Mean Matching}
\defabbrcmd{pmm}
\newabbreviation{vdm}{VDM}{Variational Diffusion Model}
\defabbrcmds{vdm}
\newabbreviation{dm}{DM}{diffusion model}
\defabbrcmds{dm}
\newabbreviation{fm}{FM}{flow matching}
\defabbrcmds{fm}
\newabbreviation{vit}{ViT}{Vision Transformer}
\defabbrcmds{vit}
\newabbreviation{dit}{DiT}{Diffusion Transformer}
\defabbrcmds{dit}
\newcommand{\unet}{U-Net}
\newcommand{\unets}{U-Nets}
\newabbreviation{cdf}{CDF}{cumulative distribution function}
\defabbrcmd{cdf}
\newabbreviation{smc}{SMC}{Sequential Monte Carlo}
\defabbrcmd{smc}
\newabbreviation{sde}{SDE}{stochastic differential equation}
\defabbrcmds{sde}
\newabbreviation{snr}{SNR}{signal-to-noise ratio}
\defabbrcmd{snr}
\newabbreviation{fid}{FID}{Fréchet inception distance}
\defabbrcmd{fid}
\newcommand{\snrsym}{\nu}

\title{Generative Modeling with Bayesian Sample Inference}

\author{\name Marten Lienen \email m.lienen@tum.de \\
  \addr School of Computation, Information and Technology \& Munich Data Science Institute\\
  Technical University of Munich
  \AND
  \name Marcel Kollovieh \email m.kollovieh@tum.de \\
  \addr School of Computation, Information and Technology \& Munich Data Science Institute\\
  Technical University of Munich
  \AND
  \name Stephan Günnemann \email s.guennemann@tum.de \\
  \addr School of Computation, Information and Technology \& Munich Data Science Institute\\
  Technical University of Munich}

\def\month{08}  
\def\year{2026} 
\def\openreview{\url{https://openreview.net/forum?id=n8qQPnalVW}} 

\begin{document}

\maketitle

\begin{abstract}
  We present a novel view of diffusion-like generative modeling from the perspective of iterative Gaussian posterior inference.
  By treating the generated sample as an unknown variable, we formulate the sampling process in the language of Bayesian probability: at each step, a model predicts the unknown sample from our current belief state and we compute a posterior belief from that prediction.
  Based on this formulation, we propose the generative model Bayesian Sample Inference (BSI).
  In addition to a rigorous theoretical analysis, we show that our perspective includes Bayesian Flow Networks (BFNs) and that BSI and BFNs represent the two opposite ends of a range of isotropic hyper-priors over the initial belief state.
  In our experiments, we demonstrate that BSI improves sample quality over both BFNs and the closely related Variational Diffusion Models on ImageNet32 and ImageNet64, while achieving equivalent log-likelihoods on both datasets.
\end{abstract}

\extrafootertext{Find a self-contained implementation and an interactive notebook at \href{https://github.com/martenlienen/bsi}{github.com/martenlienen/bsi}.}

\section{Introduction}\label{sec:introduction}

The field of deep learning has produced a multitude of generative models over the years \citep{harshvardhan2020comprehensive}.
Variational autoencoders, for example, learn the data distribution by compressing data into a lower-dimensional representation \citep{kingma2013autoencoding}.
Normalizing flows learn to map between a prior and the data distribution via invertible transformations, enabling exact likelihood computation \citep{rezende2015variational}.
Generative adversarial networks generate samples by pitting two models against each other such that one proposes artificial data samples while the other tries to distinguish real from generated samples \citep{goodfellow2014generative}.
Recently, \dms{} have become a cornerstone of generative modeling \citep{sohl-dickstein2015deep, ho2020denoising}.
They define a multi-step forward process that gradually adds noise to the data, turning it into pure noise.
Then, a model is trained to reverse this process, enabling the generation of new data samples by starting from noise and iteratively denoising.

In this work, we take a Bayesian viewpoint of sample generation and propose our generative model \bsi{} based on it.
Imagine that a sample $\vx$ from the data distribution $\pr(\vx)$ is fixed but unknown to us; however, we can receive noisy measurements $\vy_{i} \sim \N(\vx, \alpha_{i}\inv)$ of it.
Then, we can infer the unknown $\vx$ by combining the information in these measurements.
To be more precise, we start with a broad belief $\N(\vx \mid \vmu_{0}, \lambda_{0}\inv)$ about $\vx$ in the form of a Normal distribution with low precision $\lambda_{0}$, i.e.\ high variance, that encompasses the entire data distribution.
Then, we can take a first noisy measurement $\vy_{1}$ and form a posterior belief $\pr(\vx \mid \vy_{1})$ about the sample, which will be a little more precise and a little more correct.
Iterating this process allows us to refine our estimate $\pr(\vx \mid \vy_{1}, \ldots, \vy_{\nrounds})$ to any desired level of precision.

We transform this inference process into a generative model by introducing a prediction model $\model$ that estimates $\vx$ from our current Gaussian belief about it.
Since the true $\vx$ is unknown at generation time, we substitute it with an estimate $\vxhat = \model(\vmu_{i}, \lambda_{i})$ and sample $\vy_{i+1} \sim \N(\vxhat, \alpha_{i+1}\inv)$ instead.
Maximizing an \elbo{} for the likelihood that this simple process assigns to the training data trains $\model$ to reconstruct the true $\vx$ from uncertain belief states $(\vmu_{i}, \lambda_{i})$ about it.
Consequently, the noisy measurements $\vy_{i}$ of predicted samples $\vxhat$ become indistinguishable from those of real samples $\vx$, and our generative process converges toward producing new samples from the data distribution.

\begin{figure*}[t]
	\centering
	\input{figures/illustration.pgf}
	\caption{We view generation as the problem of inferring the identity of an unknown sample $\vx$ from noisy observations.\ \textbf{1}.~To begin, our belief about $\vx$ is so broad as to cover the complete data distribution.\ \textbf{2}.~We use a model~$\model$ to guess which $\vx$ likely corresponds to the information we have collected so far.\ \textbf{3}.~Now, we pretend that $\mathcolor[rgb]{0.996078431372549,0.7019607843137254,0.03137254901960784}{\vxhat}$ is the true $\vx$ and take a noisy measurement $\mathcolor[rgb]{0.8274509803921568,0.28627450980392155,0.3058823529411765}{\vy}$.\ \textbf{4}.~We form the posterior belief about $\vx$ to incorporate the information contained in $\mathcolor[rgb]{0.8274509803921568,0.28627450980392155,0.3058823529411765}{\vy}$.\ \textbf{5}.~Repeat until we have identified a new sample with sufficient precision $\lambda_{i}$.}\label{fig:illustration}
\end{figure*}

Our key \textbf{contributions} can be summarized as follows.
\begin{itemize}
	\item We present a Bayesian perspective on iterative sample generation and propose \bsi{}, a generative model based on posterior inference from noisy predictions.
	\item We derive an \elbo{} to enable effective likelihood optimization and show how we can reduce the variance of the training loss with importance sampling.
	\item We show that \bsi{} and \bfns{} \citep{graves2023bayesian} occupy the two ends of the range of reasonable isotropic hyper-priors over the initial belief, i.e.\ hyper-priors whose precision is at least the precision $\lambda_{0}$ that we ascribe to the initial belief sampled from them, with \bsi{} at the least informative end and \bfns{} at the deterministic one.
	\item Further, we compare \bsi{} in detail to \vdms{} \citep{kingma2023variational} and analyze its relationship to \dms{} and stochastic interpolants \citep{albergo2025stochastic}, showing that \bsi{} corresponds to a \dm{} with a non-Markovian forward process.
	\item Finally, we describe our model design and demonstrate empirically that \bsi{} improves sample quality over both \vdm{} and \bfn{} on ImageNet32 and ImageNet64 at equivalent log-likelihoods.
\end{itemize}

\paragraph{Notation}
We parametrize Normal distributions either with a variance $\sigma^{2}$ as $\N(\vmu, \sigma^{2}\mI)$ or with a precision $\lambda = \nicefrac{1}{\sigma^{2}}$ as $\Np(\vmu, \lambda\mI)$.
Since all Normal distributions in this work are isotropic, we shorten these to $\N(\vmu, \sigma^{2})$ and $\Np(\vmu, \lambda)$.
$[n]$~is the set of integers $1, \ldots, n$ and $\Rplus$ refers to the non-negative reals.

\section{Sample Discovery through Iterative Measurement}\label{sec:sample-discovery}

Consider a sample $\vx \in \R^{\ndim}$ that is unknown to us, but we can access noisy measurements $\vy_{i}\sim \Np(\vx, \alpha_i)$ of it.
Then we can infer $\vx$ from the sequence of measurements $\vy_{i}$ through Bayesian inference.
We start with a broad initial belief $\Np(\vmu_{0}, \lambda_{0})$ about $\vx$ and update it with the information contained in $\vy_{i}$ per the following lemma.
\begin{lemma}[Posterior Update]\label{thm:posterior}
	Let $\vx, \vmu \in \R^{\ndim}$ and $\lambda \in \Rplus$ such that $\vx$ is latent and $\pr(\vx) = \Np(\vx \mid \vmu, \lambda)$ is a prior on $\vx$; and $\vy \sim \Np(\vx, \alpha)$ where $\alpha \in \Rplus$.
	Then the posterior is $\pr(\vx \mid \vy) = \Np(\vx \mid \vmu', \lambda')$ with
	\begin{equation}
		\lambda' = \lambda + \alpha \quad \text{and} \quad \vmu' = \nicefrac{1}{\lambda'} \left[ \lambda\vmu + \alpha\vy \right]. \label{eq:posterior-update}
	\end{equation}
\end{lemma}
\begin{proof}
	See \citep[Section 4.4.1]{murphy2012machine}.
\end{proof}
We can now iterate over the noisy measurements and update our belief until $\pr(\vx \mid \vy_{1}, \ldots, \vy_{\nrounds}) = \Np(\vmu_{\nrounds}, \lambda_{\nrounds})$ identifies $\vx$ with sufficient precision.
Sufficiency depends on the application but could be defined, for example in the case of images, such that most of the probability mass for each dimension of an image $\vx$ is contained within a single color intensity bin of width $\nicefrac{1}{256}$ for 8-bit color.
Note that, at each step $i$, all information contained in $\vy_{1}, \ldots, \vy_{i}$ is captured in the current $\vmu_{i}$.

\section{Sample Generation with Posterior Inference}\label{sec:framework}

Now, we turn the procedure in \cref{sec:sample-discovery} into a generative model.
We begin with an initial belief $(\vmu_{0}, \lambda_{0})$ about the sample $\vx$ which we will generate in the end, with $\vmu_{0}$ sampled from a suitable prior distribution $\pr(\vmu_{0})$ and $\lambda_{0}$ fixed.
By construction, $\vx$ is unknown a priori, so we cannot measure it, but we can estimate it from the information we have gathered so far.

\begin{wrapfigure}[13]{r}{.47\textwidth}
	\vspace{-.7\intextsep}
	\algcaption{Sampling with posterior inference}
	\label{alg:generate}
	\begin{algorithmic}[1]
		\INPUT Initial precision $\lambda_{0}$,\\precision schedule $\alpha_{i}$ for $i \in [\nrounds]$
		\OUTPUT Sample $\vxhat^{*}$
		\STATE Initialize belief $(\vmu_{0}, \lambda_{0})$ with $\vmu_{0} \sim \pr(\vmu_{0})$
		\FOR{$i = 1, 2, \ldots, \nrounds$}
		\STATE $\vxhat_{i-1} = \model(\vmu_{i - 1}, \lambda_{i - 1})$
		\STATE $\vy_{i} \sim \Np(\vxhat_{i-1}, \alpha_{i})$
		\STATE Update belief $\pr(\vx \mid \vy_{1}, \ldots, \vy_{i}) = \Np(\vmu_{i}, \lambda_{i})$:
		\STATE \quad $\lambda_{i} = \lambda_{i-1} + \alpha_{i}$
		\STATE \quad $\vmu_{i} = \nicefrac{1}{\lambda_{i}} [\lambda_{i-1} \vmu_{i-1} + \alpha_{i}\vy_{i}]$
		\ENDFOR
		\STATE Return $\vxhat^{*} = \model(\vmu_{\nrounds}, \lambda_{\nrounds})$
	\end{algorithmic}
\end{wrapfigure}
Let $\model : \R^{\ndim} \times \Rplus \to \R^{\ndim}$ be a learned model with parameters $\vtheta$ that estimates which unknown sample $\vx$ we are observing based on our current belief $(\vmu_{i}, \lambda_{i})$.
We estimate $\vx$ as $\vxhat_{i-1} = \model(\vmu_{i-1}, \lambda_{i-1})$ and sample a noisy measurement $\vy_{i} \sim \Np(\vxhat_{i-1}, \alpha_{i})$ of $\vxhat_{i-1}$ in place of $\vx$ with precision $\alpha_{i}$.
Then, we can update our belief with $\vy_{i}$ and \cref{thm:posterior} to the posterior $(\vmu_{i}, \lambda_{i})$.
Now, we alternate between these two steps, i.e.\ predicting and taking a noisy measurement followed by updating our current belief, until the posterior precision $\lambda_{i}$ is sufficient.
Finally, we return $\vxhat^{*} = \model(\vmu_{\nrounds}, \lambda_{\nrounds})$ as our generated sample.
See \cref{alg:generate} for a formal description and \cref{fig:illustration} for a visual explanation.

Since the posterior precision $\lambda_{i}$ does not depend on the generated sample $\vxhat_{i}$, we can choose the number of measurement rounds $\nrounds$ and precision schedule $\alpha_{i}$ a priori such that $\lambda_{\nrounds}$ will always be sufficiently large.

We have collected the proofs of all formal statements in this section in \cref{sec:proofs}.

\subsection{Evidence Lower Bound}\label{sec:elbo}

By interpreting the generative process above as a hierarchical latent variable model, we derive an \elbo{} \citep{kingma2013autoencoding}, i.e.\ a lower bound on $\log\pr(\vx)$ assigned to a data point by our model.
The \elbo{} will then serve as a natural training target for $\model$ to ensure that true data samples have high likelihood under our model.

We form our hierarchy out of the sequence of belief means $\{\vmu_{i}\}$, giving us
\begin{equation}
	\pr(\vx) = \E_{\pr(\vmu_{0}) \cdot \pr(\vmu_{1} \mid \vmu_{0}) \cdots \pr(\vmu_{\nrounds} \mid \vmu_{\nrounds - 1})}[\pr(\vx \mid \vmu_{\nrounds})].
\end{equation}
The precisions $\{\lambda_{i}\}$ are not included as latent variables, because they do not depend on $\vx$.
With this hierarchy, we can derive the following \elbo{}.

\begin{restatable}{theorem}{thmelbo}\label{thm:elbo}
	Let $\vx \in \R^{\ndim}$ and $\alphaR, \alpha_{i} \in \Rplus, i \in [\nrounds]$.
	Then the log-likelihood of $\vx$ is lower-bounded as
	\begin{equation}
		\log \pr(\vx) \ge -\lrecon - \lmeasurek
	\end{equation}
	by a reconstruction term $\lrecon$ and a measurement term $\lmeasurek$,
	\begin{equation}
		\lrecon = \E_{\qr(\vmu_{\nrounds} \mid \vx, \lambda_{\nrounds})}\big[{\shortminus}\negthinspace\log\Np(\vx \mid \vxhat_{k}, \alphaR)\big] \quad \text{and} \quad \lmeasurek = \frac{\nrounds}{2} \E_{\substack{i \sim \mathcal{U}(0, k - 1) \\\qr(\vmu_{i} \mid \vx, \lambda_{i})}} \Big[ \alpha_{i+1} \|\vx - \vxhat_{i}\|_{2}^{2} \Big]
	\end{equation}
	where
	\begin{equation}
		\qr(\vmu_{i} \mid \vx, \lambda_{i}) = \E_{\pr(\vmu_{0})} \big[\pr(\vmu_{i} \mid \vmu_{0}, \vx, \lambda_{i})\big], \quad \vxhat_{i} = \model(\vmu_{i}, \lambda_{i}) \quad\text{and}\quad \lambda_{i} =  \lambda_{0} + \sum_{j=1}^{i} \alpha_{j}.
	\end{equation}
\end{restatable}

The measurement term $\lmeasurek$ corresponds to the noisy measurement and update loop in \cref{alg:generate} and $\lrecon$ to the final computation of the sample $\vxhat^{*}$.
$\qr(\vmu_{i} \mid \vx, \lambda_{i})$ is the distribution of our belief $(\vmu_{i}, \lambda_{i})$ about the unknown sample $\vx$ after $i$ steps if we had measured the true $\vx$ instead of the predictions $\vxhat_{0}, \ldots, \vxhat_{i - 1}$.
$\pr(\vmu_{i} \mid \vmu_{0}, \vx, \lambda_{i})$ is the marginal distribution of possible posterior beliefs $(\vmu_{i}, \lambda_{i})$ with posterior precision $\lambda_{i}$ reachable from an initial belief $(\vmu_{0}, \lambda_{0})$.
Equivalently, $\pr(\vmu_{i} \mid \vmu_{0}, \vx, \lambda_{i})$ is the distribution of beliefs $(\vmu_{i}, \lambda_{i})$ after updating our initial belief $(\vmu_{0}, \lambda_{0})$ with a single measurement of $\vx$ with \cref{thm:posterior} -- marginalized over all possible noisy measurements $\vy$ at precision $\alpha = \lambda_{i} - \lambda_{0}$.

On closer examination, we see that $\lrecon$, measuring how accurately we can reconstruct $\vx$ at the end, only depends on the total precision $\lambda_{k}$ that we accumulated in the first phase of the algorithm.
However, $\lmeasurek$ depends both on the number of rounds $\nrounds$ and the precision schedule $\alpha_{i}$.
We can derive an \elbo{} that is independent of $\nrounds$ and $\alpha_{i}$ by considering the limit as $\nrounds \to \infty$ and refining the precision schedule $\{\alpha_{i}\}_{i = 1}^{\nrounds}$ into smaller and smaller steps while keeping the total precision $\alphaM = \sum_{i = 1}^{\nrounds} \alpha_{i}$ constant.

\begin{restatable}{theorem}{thminfiniteelbo}
	\label{thm:infinite-elbo}
	Let $\alphaR, \alphaM \in \Rplus$.
	For any sequence of precision schedules $\alpha_{k,i}$ for $k \in \sN, i \in [k]$ such that $\sum_{i = 1}^{k} \alpha_{k, i} = \alphaM$ and the sequence of functions $[k] \to \Rplus: i \mapsto \alpha_{k,i}$ converges uniformly to $0$, we can take the limit of \cref{thm:elbo} as $k \to \infty$ to get
	\begin{equation}
		\lrecon = \E_{\qr(\vmu_{\lambdaM} \mid \vx, \lambdaM)}\big[{\shortminus}\negthinspace\log\Np(\vx \mid \vxhat_{\lambdaM}, \alphaR)\big] \quad \text{and} \quad \lmeasureinf = \frac{\alphaM}{2} \E_{\substack{\lambda\sim\mathcal{U}(\lambda_{0}, \lambdaM) \\ \qr(\vmu_{\lambda} \mid \vx, \lambda)}} \big[\|\vx - \vxhat_{\lambda}\|_{2}^{2}\big] \label{eq:infinite-elbo}
	\end{equation}
	where $\qr(\vmu_{\lambda} \mid \vx, \lambda) = \E_{\pr(\vmu_{0})} \big[\pr(\vmu_{\lambda} \mid \vmu_{0}, \vx, \lambda)\big]$, $\lambdaM = \lambda_{0} + \alphaM$ and $\vxhat_{\lambda} = \model(\vmu_{\lambda}, \lambda)$.
\end{restatable}

As long as our model is more accurate in reconstructing $\vx$ from more precise measurements, a reasonable assumption, \cref{thm:infinite-elbo} is a tighter bound on the log-likelihood than \cref{thm:elbo}.
To see this, we rewrite $\lmeasureinf$ in terms of the expected squared error at belief precision $\lambda$
\begin{equation}
	h(\lambda) = \E_{\qr(\vmu_{\lambda} \mid \vx, \lambda)} \|\vx - \vxhat_{\lambda}\|_{2}^{2} \label{eq:h}
\end{equation}
as
\begin{equation}
	\lmeasureinf = \frac{\alphaM}{2} \E_{\substack{\lambda\sim\mathcal{U}(\lambda_{0}, \lambdaM)}} [h(\lambda)]
\end{equation}
for which we have the following result.

\begin{restatable}{lemma}{thmelbotighter}
	\label{thm:elbo-tighter}
	Let $h$ be strictly decreasing.
	Then $\lmeasureinf < \lmeasurek$ for any $k$ and any precision schedule $\{\alpha_{i}\}_{i = 1}^{k}$.
	Moreover, the bound tightens monotonically under refinement of the schedule: if the partition $\{\lambda_{j}'\}_{j = 0}^{k'}$ induced by a schedule $\{\alpha_{j}'\}_{j = 1}^{k'}$ contains the partition $\{\lambda_{i}\}_{i = 0}^{k}$ induced by $\{\alpha_{i}\}$ as a strict subset, then
	\begin{equation}
		\lmeasureinf < \lmeasurekprime < \lmeasurek.
	\end{equation}
\end{restatable}

\subsection{Prior Distribution}\label{sec:prior-distribution}

It remains to specify the prior distribution $\pr(\vmu_{0})$ over the initial belief state.
We consider isotropic priors of the form $\pr(\vmu_{0}) = \Np(\vzero, \gamma_{0})$, for which we have the following result for the encoding distribution $\qr(\vmu_{\lambda} \mid \vx, \lambda)$ in \cref{thm:elbo,thm:infinite-elbo}.

\begin{restatable}{lemma}{thmelboclosedform}\label{thm:elbo-encoder-closed-form}
	Let $\lambda_{0}, \gamma_{0} \in \Rplus$, $\pr(\vmu_{0}) = \Np(\vzero, \gamma_{0})$ and $\lambda \ge \lambda_{0}$.
	Then
	\begin{equation}
		\qr(\vmu_{\lambda} \mid \vx, \lambda) = \Np\bigg( \frac{\lambda - \lambda_{0}}{\lambda}\,\vx, \frac{\lambda^{2}}{\lambda - \lambda_{0} + \nicefrac{\lambda_{0}^{2}}{\gamma_{0}}} \bigg).
	\end{equation}
\end{restatable}

How should we choose the hyper-prior precision $\gamma_{0}$?
If $\gamma_{0}$ were less than $\lambda_{0}$, we would be unreasonably confident in our initial belief: we would believe that our initial belief had precision $\lambda_{0}$, when we know that it can have at most $\gamma_{0} < \lambda_{0}$.
This constrains $\gamma_{0}$ from below and we deduce that the reasonable range for the belief-prior precision $\gamma_{0}$ is $[\lambda_{0}, \infty]$.

The two ends of this range are the only choices for which the encoder precision in \cref{thm:elbo-encoder-closed-form} simplifies.
At $\gamma_{0} = \lambda_{0}$, the hyper-prior is just as precise as the initial belief claims to be, and we get the following result.

\begin{restatable}{corollary}{thmbsielboclosedform}\label{thm:bsi-elbo-encoder-closed-form}
	Let $\lambda_{0} \in \Rplus$, $\pr(\vmu_{0}) = \Np(\vzero, \lambda_{0})$ and $\lambda \ge \lambda_{0}$.
	Then
	\begin{equation}
		\qr(\vmu_{\lambda} \mid \vx, \lambda) = \Np\bigg(\frac{\lambda - \lambda_{0}}{\lambda}\,\vx, \lambda\bigg).
	\end{equation}
\end{restatable}

With this choice, the encoder precision reduces to the belief precision $\lambda$ itself.
\emph{We refer to the resulting model as \bsi{}.}

At the opposite end of the range, $\gamma_{0} = \infty$, the prior collapses to the deterministic $\vmu_{0} = \vzero$, so that sampling always starts from the same deterministic belief and the encoder precision becomes $\lambda^{2}{(\lambda - \lambda_{0})}\inv$.
We show in \cref{sec:bfns} that this choice recovers \bfns{} if we additionally fix the initial precision to $\lambda_{0} = 1$ as they do.
Thus, \bsi{} and \bfns{} are closely related and differ in their choice of hyper-prior for the initial belief: \bsi{} uses the least informative prior in this range, while \bfns{} choose the most restricted one.

\subsection{Variance Reduction}\label{sec:variance-reduction}

The squared distance $\|\vx - \vxhat_{\lambda}\|_{2}^{2}$ in $\lmeasureinf$ will necessarily vary significantly across the range of $\lambda$ with large values for small $\lambda$ where $\qr(\vmu_{\lambda} \mid \vx, \lambda) \approx \pr(\vmu_{0})$ and small values for large $\lambda$ when $\vmu_{\lambda} \approx \vx$.
We can reduce the variance of \mc{} estimates of $\lmeasureinf$ for \elbo{} evaluation or gradient computation in training with importance sampling with a suitable proposal distribution $\pr(\lambda)$.

\begin{restatable}{corollary}{thmimportancesampling}
	\label{thm:importance-sampling}
	Let $\pr(\lambda)$ be a probability distribution with support $[\lambda_{0}, \lambdaM]$.
	Then we have
	\begin{equation}
		\lmeasureinf = \frac{1}{2} \E_{\substack{\lambda\sim\pr(\lambda) \\ \qr(\vmu_{\lambda} \mid \vx, \lambda)}} \bigg[ \frac{1}{\pr(\lambda)} \|\vx - \vxhat_{\lambda}\|_{2}^{2} \bigg].
	\end{equation}
\end{restatable}

We can further rewrite $\lmeasureinf$ as
\begin{equation}
	\lmeasureinf = \frac{1}{2} \E_{\lambda\sim\pr(\lambda)} \bigg[ \frac{h(\lambda)}{\pr(\lambda)} \bigg] \label{eq:is-lm}
\end{equation}
with $h$ as defined in \cref{eq:h}.
To minimize the variance of \mc{} estimates of $\lmeasureinf$, we want to bring $\nicefrac{h(\lambda)}{\pr(\lambda)}$ as close to a constant as possible.
If it were actually constant, the variance of the \mc{} estimate would be zero.

Let's begin by examining $h$ more closely.
If we approximate $\model$ as $\model(\vmu, \lambda) = \vmu$ and assume that $\vx$ is normalized to zero mean and unit variance, we get the closed form
\begin{equation}
	\E_{\vx}[h(\lambda)] \propto \frac{\lambda_{0}^{2}}{\lambda^{2}} + \frac{1}{\lambda}. \label{eq:expected-h}
\end{equation}
While $\model(\vmu, \lambda) = \vmu$ might seem a crude approximation at first, it is not too far off for large $\lambda$ where the model just needs to predict a small correction to its input.

\cref{eq:expected-h} suggests that we should choose $\pr(\lambda) \propto \nicefrac{\lambda_{0}^{2}}{\lambda^{2}} + \nicefrac{1}{\lambda}$ to minimize the variance of \mc{} estimates.
While evaluating $\pr(\lambda)$ is simple enough, we would need to invert its \cdf{} numerically to sample from it.
Instead, we recognize that $\nicefrac{1}{\lambda}$ dominates $\nicefrac{\lambda_{0}^{2}}{\lambda^{2}}$ except for the smallest $\lambda$ and choose $\pr(\lambda) \propto \nicefrac{1}{\lambda}$, i.e.\ a standard $\LogUniform(\lambda_{0}, \lambdaM)$ distribution.

\subsection{Training \& Sampling}\label{sec:algorithms}

\begin{wrapfigure}[19]{r}{.45\textwidth}
	\vspace{-.7\intextsep}
	\algcaption{Estimating the BSI training loss}
	\label{alg:loss}
	\begin{algorithmic}[1]
		\INPUT Data sample $\vx$
		\OUTPUT Monte Carlo estimate of $2\lmeasureinf$
		\STATE Sample $t \sim \mathcal{U}(0, 1)$, $\vepsilon \sim \N(\vzero, \mI)$
		\STATE $\lambda = \exp\big((\log\lambdaM - \log\lambda_{0}) \cdot t + \log(\lambda_{0})\big)$
		\STATE $\vmu_{\lambda} = \nicefrac{(\lambda - \lambda_{0})}{\lambda}\,\vx + \sqrt{\nicefrac{1}{\lambda}}\,\vepsilon$
		\STATE Return $(\log\lambdaM - \log\lambda_{0})\,\lambda \cdot \|\vx - \model(\vmu_{\lambda}, \lambda)\|_{2}^{2}$
	\end{algorithmic}
	\algcaption{Sampling with BSI}
	\label{alg:generate-specific}
	\begin{algorithmic}[1]
		\INPUT Initial precision $\lambda_{0}$,\\precision schedule $\alpha_{i}$ for $i \in [\nrounds]$
		\OUTPUT Sample $\vxhat^{*}$
		\STATE Sample $\vepsilon_{i} \sim \N(\vzero, \mI), i = 0, \ldots, \nrounds$
		\STATE $\vmu_{0} = \sqrt{\nicefrac{1}{\lambda_{0}}}\,\vepsilon_{0}$
		\FOR{$i = 1, 2, \ldots, \nrounds$}
		\STATE $\vxhat_{i-1} = \model(\vmu_{i - 1}, \lambda_{i - 1})$
		\STATE $\lambda_{i} = \lambda_{i - 1} + \alpha_{i}$
		\STATE $\vmu_{i} = \lambda_{i}\inv \Big(\lambda_{i-1} \vmu_{i-1} + \alpha_{i}\big(\vxhat_{i-1} + \sqrt{\nicefrac{1}{\alpha_{i}}}\vepsilon_{i}\big)\Big)$
		\ENDFOR
		\STATE Return $\vxhat^{*} = \model(\vmu_{\nrounds}, \lambda_{\nrounds})$
	\end{algorithmic}
\end{wrapfigure}

We train our model with the \elbo{} from \cref{thm:infinite-elbo} by optimizing $2\lmeasureinf / \ndim$.
We do not optimize $\lrecon$ directly as its magnitude is negligible for sufficiently large $\alphaM$ -- by then the belief is concentrated enough that the final reconstruction is all but determined, see \cref{sec:hyperparameters} -- and it is structurally similar to $\lmeasureinf$, i.e.\ both amount to a squared distance.
\cref{alg:loss} shows the resulting algorithm with our belief prior $\pr(\vmu_{0})$ and proposal distribution $\pr(\lambda)$.
Similarly, \cref{alg:generate-specific} implements the abstract \cref{alg:generate} with our belief prior.

\section{Discussion}\label{sec:discussion}

We are aware of two generative models that are closely related to \bsi{}: \bfns{} \citep{graves2023bayesian} and \vdms{} \citep{kingma2023variational}.
\bfns{} are generative models motivated from an information-theoretic perspective with a sender and a receiver communicating about the sample.
As we show in \cref{sec:bfns}, \bfns{} can be understood from the theoretical perspective in \cref{sec:framework}, where they occupy the end of the belief-prior range opposite to \bsi{}: they correspond to $\gamma_{0} = \infty$ and $\lambda_{0} = 1$, meaning that sampling always starts from the deterministic belief $(\vmu_{0}, \lambda_{0}) = (\vzero, 1)$.
In contrast, for \bsi{} we choose $\gamma_{0} = \lambda_{0}$, i.e.\ the noise in the initial belief corresponds to our confidence in it, and leave $\lambda_{0}$ as a hyperparameter, which we investigate in \cref{sec:experiments}.
\vdms{} are a type of \dm{} that have shown excellent performance in likelihood-based modeling.
They are similar to \bsi{} insofar as they specify the distribution of latent variables directly rather than defining a Markovian noising process as classical \dms{} do.

\begin{table*}[t]
	\centering
	\caption{Central structures of \vdm{}, \bfn{} and \bsi{}. To improve comparability, we parametrize \vdm{} in terms of the \snr{}~$\snrsym$. \bfn{} and \bsi{} are parametrized with the belief precision~$\lambda$ as introduced in \cref{sec:framework}. $\vepsilon_{i} \sim \N(\vzero, \mI)$ is sampling noise.}\label{tab:distributions}
	\resizebox{\textwidth}{!}{
		\begin{tabular}{rccc}
			\toprule
			Model & \multicolumn{1}{c}{ELBO Encoder $\qr(\vpsi \mid \vx, \omega)$}                                                     & Latent Prior                             & Update Step for Sampling                                                                                                                                                                                                             \\
			\midrule
			VDM   & $\qr(\vz \mid \vx, \snrsym) = \Np\Big(\sqrt{\frac{\snrsym}{1 + \snrsym}}\,\vx, 1 + \snrsym\Big)$                   & $\vz_{T} \sim \Np(\vzero, 1)$            & $\vz_{i} = \frac{\sqrt{\snrsym_{i + 1}(1 + \snrsym_{i + 1})}\,\vz_{i + 1} + (\snrsym_{i} - \snrsym_{i + 1})\big(\vxhat_{i} + \sqrt{\frac{1}{\snrsym_{i} - \snrsym_{i + 1}}}\vepsilon_{i}\big)}{\sqrt{\snrsym_{i}(1 + \snrsym_{i})}}$ \\
			\midrule
			BFN   & $\qr(\vmu \mid \vx, \lambda) = \Np(\nicefrac{(\lambda - 1)}{\lambda}\,\vx, \nicefrac{\lambda^{2}}{(\lambda - 1)})$ & $\vmu_{0} = \vzero$                      & \multirow{2}{*}{$\vmu_{i} = \frac{\lambda_{i-1} \vmu_{i-1} + \alpha_{i}\big(\vxhat_{i-1} + \sqrt{\frac{1}{\alpha_{i}}}\vepsilon_{i}\big)}{\lambda_{i - 1} + \alpha_{i}}$}                                                            \\
			\cmidrule{1-3}
			BSI   & $\qr(\vmu \mid \vx, \lambda) = \Np(\nicefrac{(\lambda - \lambda_{0})}{\lambda}\,\vx, \lambda)$                     & $\vmu_{0} \sim \Np(\vzero, \lambda_{0})$ &                                                                                                                                                                                                                                      \\
			\bottomrule
		\end{tabular}}
\end{table*}
\begin{figure*}[t]
	\centering
	\includegraphics{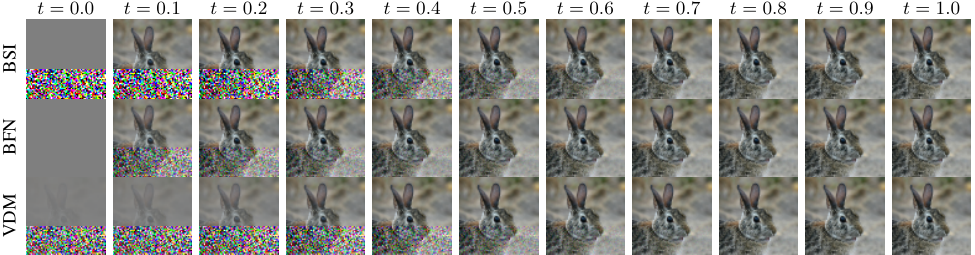}
	\caption{\elbo{} encoders $\qr$, i.e.\ training inputs, of \bsi{}, \bfn{} and \vdm{}. $t$ parametrizes the precision levels by the respective model's precision schedule with $t = 0$ being pure noise, ideally, and $t = 1$ almost equaling the data. Top half shows the mean of $\qr$ and bottom half a sample. Mean $\vzero$ is gray because all models rescale the data to $[-1, 1]$.\ \bfns{} apply little noise overall and reach a deterministic state at $t = 0$. For \vdm{}, significant information about the sample is preserved in the structure of the mean at the highest noise level. In contrast, \bsi{} converges to its latent prior distribution.}\label{fig:noise-levels}
\end{figure*}

All three models admit an \elbo{} of the form
\begin{equation}
	-\log\pr(\vx) \le \lrecon + \frac{\bar{\omega} - \underbar{\omega}}{2} \E_{\substack{\omega\sim\mathcal{U}(\underbar{\omega}, \bar{\omega}) \\ \qr(\vpsi_{\omega} \mid \vx, \omega)}} \big[\|\vx - \vxhat_{\omega}\|_{2}^{2}\big]
\end{equation}
for a set of latent variables $\vpsi$ at precision levels $\omega$ between $\underbar{\omega}$ and $\bar{\omega}$.
For \bsi{} and \bfn{}, the precision level $\omega$ is the belief precision $\lambda$ between $\lambda_{0}$ and $\lambdaM$ and $\vpsi_{\omega} = \vmu_{\lambda}$.
For \vdm{}, the latent variables $\vpsi$ are called $\vz$ and the precision level $\omega$ is the \snr{} $\snrsym$ between $e^{-5}$ and $e^{13.3}$.

Despite this shared \elbo{} form, the models vary significantly in the specifics.
\cref{tab:distributions} lists the encoding distribution $\qr(\vpsi \mid \vx, \omega)$ for each model, their prior, from which they begin the sampling process, and the update step that the models iterate during sampling.
First, we see that \vdm{} starts sampling from a standard Normal vector and \bfn{} from the deterministic $\vzero$.
Only \bsi{} allows starting the sampling process from an initial belief with a precision $\lambda_{0}$ below $1$, i.e.\ a variance above $1$, which has been shown to improve sample diversity in consistency models \citep{song2024improved}.
Second, the update step shared between \bsi{} and \bfn{} is significantly simpler than the \vdm{} update with respect to the precision parameter and does not require evaluation in log-space for numerical stability as recommended for \vdm{} \citep{kingma2023variational}.

For the encoding distribution $\qr(\vpsi \mid \vx, \omega)$, which provides the training inputs when the models optimize their \elbo{}, we turn to \cref{fig:noise-levels}.
First, we note that \bfns{} add little noise overall due to their noise variance $\nicefrac{(\lambda - 1)}{\lambda^{2}}$ going to $0$ for both small and large $\lambda$.
Next, we notice the encoding distribution $\qr(\vpsi \mid \vx, \underbar{\omega})$ with the most noise at $t = 0$.
While it agrees exactly with the latent prior used for sampling for \bsi{} and \bfn{}, for \vdm{} it becomes approximately $\Np(0.08\vx, 1)$, which differs significantly from the standard Normal prior for sampling.
In fact, the image motif is still clearly discernible in the distribution mean for \vdm{} at its maximum noise level.
The amount of signal remaining in the mean for \bsi{} at high noise levels is counteracted by much higher noise variance, e.g.\ $15.85$ at $t = 0.1$ for \bsi{} compared to $0.96$ for \vdm{}.

\paragraph{Diffusion Models}
If we currently hold the belief $(\vmu', \lambda')$, the distribution over beliefs $(\vmu, \lambda)$ with $\lambda = \lambda' - \alpha$ that are $\alpha$ less precise is
\begin{equation}
	\pr(\vmu \mid \vmu', \vx) = \Np\Bigg( \xi\inv \bigg[ \frac{\lambda\lambda'}{\alpha}\vmu' - \lambda_{0}\vx \bigg], \xi \Bigg)
\end{equation}
for a certain precision $\xi$.
This shows that \bsi{} can be written as a \dm{} with a non-Markovian forward or ``noising'' process.
See \cref{sec:diffusion-models} for a detailed derivation of this connection.
There we also exploit that \bfns{} correspond to $\gamma_{0} = \infty$ to derive the forward process for \bfn{} and show that it is Markov, in contrast to the \bsi{} process.
Beyond \dms{}, we relate \bsi{} to stochastic interpolants in \cref{sec:stochastic-interpolants}, to Posterior Mean Matching \citep{salazar2025posterior} as another posterior-inference view of generative modeling in \cref{sec:pmm}, and to \smc{} samplers in \cref{sec:smc}.

\section{Model Design}\label{sec:model-design}

In this section, we introduce a design for the prediction model in \bsi{}.
We begin by deriving a preconditioning structure for $\model$, i.e.\ a type of model structure similar to noise prediction in \dms{}.
Then, we describe how we bring $\lambda$ into a suitable range as an input for deep learning.
Finally, we give our choice of the hyperparameters $\lambda_{0}$, $\alphaM$ and $\alphaR$ and report the model architectures we used as the backbone of $\model$.

\subsection{Preconditioning}\label{sec:preconditioning}

It has long been known in the context of \dms{} that training models to predict $\vx$ directly from a noisy input can hinder learning and limit sample quality \citep{karras2022elucidating,ho2020denoising}.
For probabilistic modeling, it is especially important that the model prediction stays close to the true sample if the input is already at a low noise level to achieve high \elbos{}.
This can be seen, for example, in \cref{thm:importance-sampling} where prediction errors for high-precision input beliefs with large $\lambda$ have a higher weight.
Instead of predicting $\vx$, \dms{} commonly either predict a variation of the noise in the model input \citep{ho2020denoising,song2021denoising} or an adaptive mixture of the noise and the true sample \citep{salimans2021progressive}.
In the end, these approaches amount to adding a skip connection to the model with specific weights.

For \bsi{}, we derive such a preconditioning structure with the adaptive-mixture approach from \citet{karras2022elucidating}.
Let $\model'$ be our neural network.
Then we define the preconditioned $\model$ as
\begin{equation}
	\model(\vmu, \lambda) = \preskip \vmu + \preout \model'(\prein\vmu, \lambda) \label{eq:preconditioned-model}
\end{equation}
and find the parameters through the conditions proposed by \citet{karras2022elucidating}.
$\prein$ and $\preout$ are chosen such that the input to $\model'$ and its training target have unit variance.
$\preskip$ is then chosen to minimize $\preout$, which minimizes the influence of prediction errors and ensures that $\model$ retains most of the signal already contained in $\vmu$ at large precisions~$\lambda$.

From these conditions, we derive
\begin{equation}
	\begin{aligned}
		\preskip = \nicefrac{(\lambda - \lambda_{0})}{\kappa}, \quad \preout = \sqrt{\nicefrac{1}{\kappa}}, \quad \prein = \sqrt{\nicefrac{\lambda}{\kappa}}\end{aligned} \label{eq:preconditioning}
\end{equation}
where $\kappa = 1 + \nicefrac{{(\lambda - \lambda_0)}^{2}}{\lambda}$ in \cref{sec:preconditioning-derivation}.
$\lambda$ is the precision of our current belief about $\vx$ and the input to $\model$.

\subsection{Precision Encoding}\label{sec:precision-encoding}

The magnitude of $\lambda$ makes it impractical as a feature for neural networks.
However, the \cdf{} $\LUcdf$ of $\pr(\lambda)$ is a natural way to scale $\lambda$ from $[\lambda_{0}, \lambdaM]$ to $[0, 1]$ as in \dms{} and \fm{} \citep{lipman2023flow}.
In practice, we use $\model(\vmu, t)$ instead of $\model(\vmu, \lambda)$ where
\begin{equation}
	t = \LUcdf(\lambda) = \frac{\log\lambda - \log\lambda_{0}}{\log(\lambdaM) - \log\lambda_{0}}.
\end{equation}
Compared to linear re-scaling, our method makes it easier for $\model$ to distinguish belief precisions in the high-noise regime.

\subsection{Hyperparameters}\label{sec:hyperparameters}

Apart from $\model$, \bsi{} has three hyperparameters, $\lambda_{0}$, $\alphaM$ and $\alphaR$.
$\lambda_{0}$ should be small enough that the initial belief covers the whole data distribution.
We have found experimentally that $\lambda_{0} = 10^{-2}$ balances likelihood and sample quality for images rescaled to $[-1, 1]$, see \cref{sec:parameter-studies}.

$\alphaM$ should be large enough that a noisy measurement at precision $\alphaM$ identifies an $\vx$, e.g.\ for images almost all probability mass of $\Np(\vx, \alphaM)$ should be contained within a single 8-bit color intensity bin.
We choose $\alphaM = 10^{6}$, which \citet{graves2023bayesian} also picked for \bfn{}.
While $\lmeasureinf$ dwarfs $\lrecon$, $\alphaR = 2\alphaM$ gives a slight edge in likelihood, empirically, as also observed by \citet{graves2023bayesian}.

For sampling, we step uniformly through the inverse \cdf{} $\LUcdf\inv$ of the importance sampling distribution $\pr(\lambda)$, i.e.\ for $k$ sampling steps, we choose the precision gain $\alpha_{i}$ in the $i$-th step as
\begin{equation}
	\alpha_{i} = \LUcdf\inv\Big( \frac{i}{k} \Big) - \LUcdf\inv\Big( \frac{i - 1}{k} \Big).
\end{equation}
This corresponds to exponential steps in $\alpha_{i}$ for a log-uniform $\pr(\lambda)$.

\subsection{Architecture}\label{sec:architecture}

After the preconditioning and mapping $\lambda$ to a $t \in [0, 1]$, there are two more steps to turn the inputs $\vmu$ and $t$ of $\model'$ into effective features for a neural network.
Regarding $t$, we convert it into a $32$-dimensional precision embedding with a sinusoidal position encoding \citep{vaswani2017attention}.

The Fourier features proposed by \citet{kingma2023variational} are an essential component to reach high likelihoods, because they help the model distinguish fine details at high belief precisions, i.e.\ for inputs that are already close to the data distribution.
They are basically a sinusoidal embedding of every dimension of $\vmu$.
In particular, we extend $\vmu$ to the vector
\begin{equation}
	\begin{pmatrix}
		\vmu & \sin(2^{i}\pi \vmu) & \cos(2^{i}\pi \vmu)
	\end{pmatrix}\ i \in \{n_{\mathrm{min}}, \ldots, n_{\mathrm{max}}\}
\end{equation}
before passing it into the neural network.
We choose $n_{\mathrm{min}} = 6$ and $n_{\mathrm{max}} = 8$, in effect increasing the dimensionality of the input to the neural network from $\ndim$ to $7\ndim$.

For the neural network itself, we use two architectures, \unets{} \citep{ronneberger2015unet} and \vits{} \citep{dosovitskiy2020image}.
We use the \unet{} configuration proposed by \citet{kingma2023variational}, which adapts the widely used configuration of \citet{ho2020denoising} for likelihood estimation.
Most notably, their version has no downsampling between layers of the \unet{}, which lets them increase the number of \unet{} levels to 32.

\vits{} are a more recent architecture inspired by the success of transformers \citep{vaswani2017attention}.
They represent images as a set of patches with a 2D position embedding and process them with global attention, in contrast to convolutional architectures like the \unet{} where communication happens primarily locally.
We opt for the \dit{} architecture \citep{peebles2023scalable}, which has been shown to improve sample quality over variants of the \unet{} model of \citet{ho2020denoising}.

\section{Experiments}\label{sec:experiments}

We evaluate \bsi{} on the ImageNet \citep{deng2009imagenet} dataset in terms of log-likelihood and sample quality and on CIFAR10 \citep{krizhevsky2009learning} in terms of log-likelihood.
While \bsi{} is not specific to images, we chose image datasets, because they are established benchmarks in the probabilistic modeling literature.
In our experiments, we compare against \bfns{} \citep{graves2023bayesian} and \vdms{} \citep{kingma2023variational}.
\bfns{} sit at the opposite end of the belief-prior range $[\lambda_{0}, \infty]$ from \bsi{} (see \cref{sec:prior-distribution}) with $\lambda_{0} = 1$ and therefore isolate the joint effect of the non-deterministic hyper-prior $\pr(\vmu_{0})$ and the initial precision $\lambda_{0}$ in \bsi{}.
\vdms{} are representatives of the diffusion family of models, specifically designed for probabilistic modeling and structurally similar to \bsi{}, as we explain in \cref{sec:discussion}.

\begin{wraptable}[12]{r}{1.75in}
	\vspace{-0.15em}
	\centering
	\caption{Log-likelihood in BPD and sample quality (FID) against the test set on ImageNet. We compute standard deviations over 3 seeds.}\label{tab:imagenet32}
	\resizebox{1.75in}{!}{
		\begin{tabular}{rcc}
			\toprule
			Model & BPD $\downarrow$                                                           & FID $\downarrow$  \\
			\midrule
			      & \multicolumn{2}{c}{\scriptsize{ImageNet32 (\SI{2}{\million} train steps)}}                     \\
			BFN   & \num{3.448 +- 0.005}                                                       & \num{11.0 +- 0.1} \\
			VDM   & \num{3.452 +- 0.006}                                                       & \num{9.9 +- 0.5}  \\
			BSI   & \num{3.448 +- 0.006}                                                       & \num{8.9 +- 0.1}  \\
			\midrule
			      & \multicolumn{2}{c}{\scriptsize{ImageNet64 (\SI{1}{\million} train steps)}}                     \\
			BFN   & \num{3.222 +- 0.006}                                                       & \num{38.2 +- 0.8} \\
			VDM   & \num{3.228 +- 0.006}                                                       & \num{35.2 +- 0.7} \\
			BSI   & \num{3.218 +- 0.004}                                                       & \num{30.1 +- 0.4} \\
			\bottomrule
		\end{tabular}
	}
\end{wraptable}
In \cref{sec:bits-per-dimension}, we describe how we compute the \elbo{}, which we derived in \cref{sec:elbo} for continuous $\vx$, on discretized images with 8-bit color channels.
\cref{sec:experiment-details} lists hyperparameters and training details and \cref{sec:generated-samples} shows some generated samples.
For our ImageNet experiments, we train all three models from scratch using the same backbone architecture and optimizer settings.
On CIFAR10, we train \bsi{} in the setting of the baselines and compare against the log-likelihoods that they report, as we detail in \cref{sec:cifar10}.

\subsection{ImageNet}\label{sec:imagenet}

For this evaluation, we train a \dit{} \citep{peebles2023scalable} in the \bfn{}, \vdm{} and \bsi{} model, respectively, on the official 32\texttimes32 and 64\texttimes64 versions of ImageNet \citep{chrabaszcz2017downsampled}.
We train each model from three seeds and evaluate the log-likelihood of the test set in \bpd{} and the sample quality in terms of \fid{} against the test set.
For the log-likelihood, we evaluate each model's \elbo{} with \num{5} samples from the respective equivalent of $\lmeasureinf$ and \num{2} samples from the respective equivalent of $\lrecon$.
For the sample quality, we draw \num{50000} \emph{unconditional} samples from each model with \num{1024} steps and then compute the \fid{} between the generated samples and the test set.
On the 32\texttimes32 resolution images, we train the \dit{}-L-2 configuration for \SI{2}{\million} steps and on the 64\texttimes64 resolution data, we train the \dit{}-L-4 configuration for \SI{1}{\million} steps.

\begin{figure*}
	\begin{minipage}[t]{1.7in}
		\begin{figure}[H]
			\centering
			\includegraphics{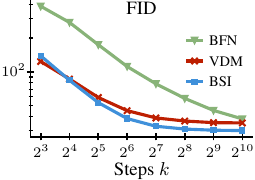}
			\caption{On ImageNet64, BSI's sample quality converges quickly and to a lower FID with increasing number of steps.}\label{fig:fid-convergence}
		\end{figure}
	\end{minipage}
	\hfill
	\begin{minipage}[t]{1.3in}
		\begin{figure}[H]
			\centering
			\includegraphics{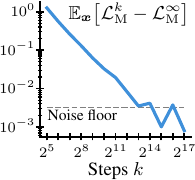}
			\caption{$\lmeasurek$ converges to $\lmeasureinf$ from above as predicted in \cref{thm:elbo-tighter}.}\label{fig:elbo-convergence}
		\end{figure}
	\end{minipage}
	\hfill
	\begin{minipage}[t]{1.65in}
		\begin{figure}[H]
			\centering
			\includegraphics{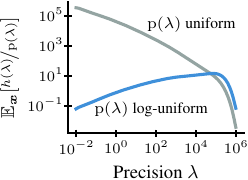}
			\caption{Our proposal distribution shrinks the range of $\nicefrac{h(\lambda)}{\pr(\lambda)}$, reducing ELBO variance.}\label{fig:importance-sampling}
		\end{figure}
	\end{minipage}
	\hfill
	\begin{minipage}[t]{1.5in}
		\begin{figure}[H]
			\centering
			\includegraphics{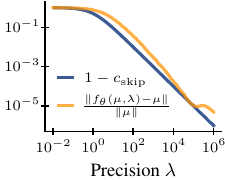}
			\caption{As $\lambda$ increases, $\vxhat = \model(\vmu, \lambda)$ and the belief $\vmu$ converge.}\label{fig:log-uniform-proposal}
		\end{figure}
	\end{minipage}
	\\
	\begin{minipage}[t]{2.2in}
		\begin{figure}[H]
			\centering
			\includegraphics{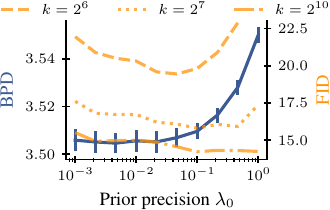}
			\caption{$\lambda_{0}$ balances likelihood and sample quality for varying sample steps $k$.}\label{fig:lambda0}
		\end{figure}
	\end{minipage}
	\hfill
	\begin{minipage}[t]{4.2in}
		\begin{figure}[H]
			\centering
			\includegraphics{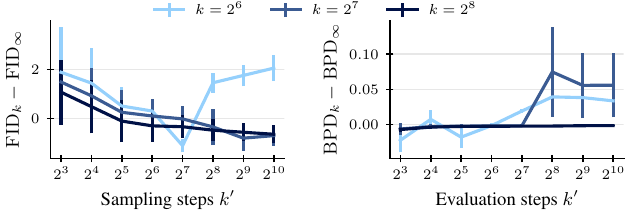}
			\caption{FID and likelihood difference between models trained on $\lmeasurek$ and $\lmeasureinf$ when evaluated for $k'$ steps.}\label{fig:train-finite-k}
		\end{figure}
	\end{minipage}
\end{figure*}

\cref{tab:imagenet32} shows that \bsi{} achieves equivalent likelihoods to \vdm{} and \bfn{} while generating higher-quality samples in terms of \fid{}.
This aligns with the result for consistency models by \citet{song2024improved} that a larger variance of the initial state -- initial belief $\vmu_{0}$ for \bsi{} -- improves sample diversity.
Ordering the models from worst to best \fid{}, we have \bfn{} first with an initial variance of $0$ ($\vmu_{0} = \vzero$), then \vdm{} with an initial variance of $1$ and finally \bsi{} with an initial variance of $\lambda_{0}\inv = 100$.
The magnitude of the FID on ImageNet64 aligns with the results reported by \citet{peebles2023scalable} after \SI{1}{\million} training steps.
Furthermore, \cref{fig:fid-convergence} shows that, from 16 steps upwards, \bsi{} achieves better sample quality than \bfn{} and \vdm{} for the same number of steps on ImageNet64.

\paragraph{ELBO Convergence}
\cref{fig:elbo-convergence} shows how the finite-step \elbo{} from \cref{thm:elbo} converges towards its infinite-step counterpart as $k \to \infty$ on the test set of ImageNet32.
For this plot, we sampled 100 precisions $\lambda$ per image for the Monte Carlo estimates of $\lmeasurek$ and $\lmeasureinf$.
The convergence trend continues right to the noise floor where the noise overshadows the signal, marked in the plot by the standard deviation of the Monte Carlo estimator for the difference between the two terms.

\subsection{CIFAR10}\label{sec:cifar10}

We train the same \unet{} architecture as \vdm{} \citep{kingma2023variational} and \bfn{} \citep{graves2023bayesian} on CIFAR10.
\cref{tab:cifar10-bpd} shows that \bsi{} achieves equivalent log-likelihoods in terms of \bpd{}.
Due to the significant number of training steps (\SI{5}{\million} and \SI{10}{\million}), we followed \citet{kingma2023variational} and \citet{graves2023bayesian} and trained only a single model per training budget on this dataset.

\paragraph{Variance Reduction}
\cref{fig:importance-sampling} verifies the effect of importance sampling with a log-uniform distribution that we propose in \cref{sec:variance-reduction}.
It reduces the range of the $\nicefrac{h(\lambda)}{\pr(\lambda)}$ term in \cref{eq:is-lm} by about 4 orders of magnitude on CIFAR10 and therefore the variance of a Monte Carlo estimate of the \elbo{}.

\subsection{Parameter Studies}\label{sec:parameter-studies}

In the following, we evaluate the impact of our modeling and parameter choices.
Unless otherwise stated, we trained each model for \SI{100}{\thousand} steps on ImageNet32 with a \dit{} architecture, evaluated the likelihood of the test set in \bpd{} with the infinite-step \elbo{} and used \num{1024} sampling steps to compute the \fid{}.
We will verify the assumptions of the log-uniform proposal distribution $\pr(\lambda)$, compare \dit{} and \unet{} model architectures, and evaluate the prior precision $\lambda_{0}$ and training on the finite-step \elbo{} $\lmeasurek$.

\paragraph{Proposal Distribution} In \cref{sec:variance-reduction}, we have chosen a log-uniform proposal distribution $\pr(\lambda)$ based on the assumption that $\model(\vmu, \lambda) \approx \vmu$.
\cref{fig:log-uniform-proposal} shows that the relative distance between $\vmu$ and $\model(\vmu, \lambda)$ falls quickly for $\lambda > 1$, when the belief $(\vmu, \lambda)$ contains enough information that the model mostly refines the belief.
Our preconditioning structure $\model(\vmu, \lambda) = \preskip \vmu + \preout \model'(\prein\vmu, \lambda)$ derived in \cref{sec:preconditioning} ensures that $\model$ retains existing information as the precision $\lambda$ grows.


\begin{wraptable}{r}{1.75in}
	\vspace{-0.15em}
	\centering
	\caption{Trained with \unet{} architecture. We compute standard deviations over 3 seeds.}\label{tab:architecture}
	\resizebox{1.75in}{!}{
		\begin{tabular}{rcc}
			\toprule
			Model & BPD                  & FID               \\
			\midrule
			BFN   & \num{3.505 +- 0.001} & \num{14.2 +- 0.4} \\
			VDM   & \num{3.527 +- 0.009} & \num{15.4 +- 1.5} \\
			BSI   & \num{3.510 +- 0.009} & \num{12.8 +- 0.6} \\
			\bottomrule
		\end{tabular}
	}
\end{wraptable}
\paragraph{Model Architecture} To ensure that the improvements in sample quality on ImageNet arise from \bsi{} as a method and not from the architecture of the underlying model, we have also trained \unets{} on ImageNet32.
\cref{tab:architecture} shows that the \unet{} exhibits the same characteristics as the \dit{} that we trained in \cref{sec:imagenet}, i.e.\ equivalent likelihoods between \bfn{}, \vdm{} and
\bsi{} with a consistent improvement in \fid{}.
We chose the \unet{} parameterization of \citet{kingma2023variational}, which is also listed in \cref{sec:experiment-details}.

\paragraph{Initial Precision} In \cref{fig:lambda0}, we evaluate the impact of the initial precision  $\lambda_{0}$ on likelihood and sample quality.
While the likelihood of test data improves with falling $\lambda_{0}$, i.e.\ increasing initial noise, the sample quality depends on the number of sampling steps.
For a large number of steps, larger values of $\lambda_{0}$ perform slightly better, but with fewer steps an intermediate $\lambda_{0}$ is preferred.
With fewer total sampling steps, decreasing $\lambda_{0}$ ensures that the sampling process still spends enough steps in the intermediate noise range, which is responsible for the generation of large-scale features in the images \citep{rissanen2022generative}.

\paragraph{Training with $\lmeasureinf$} By default, we train by optimizing the measurement loss $\lmeasureinf$ of the infinite-step \elbo{}, but in practice the model will only use finitely many steps.
\cref{fig:train-finite-k} shows that training on $\lmeasurek$ does not confer a consistent advantage in sample quality or likelihood.
This justifies training by optimizing $\lmeasureinf$ regardless of the number of steps used later and eliminates $k$ as a hyperparameter.

\section{Conclusion}\label{sec:conclusion}

We have presented a perspective on generative modeling that frames sample generation as iterative Bayesian inference and derived our generative model \bsi{} from it.
The \elbo{} we have derived for both finite steps and the infinite-step limit enables efficient training, and we reduced the training loss variance through importance sampling.
In addition, we have thoroughly discussed how \bsi{} relates to \bfn{} and \dms{} and shown that \bsi{} and \bfn{} occupy the two ends of the range of reasonable priors on the initial belief.
Our experiments have demonstrated that \bsi{} generates better samples than both \vdm{} and \bfn{} while achieving equivalent log-likelihoods on established image datasets.
Overall, \bsi{} contributes a Bayesian perspective to the landscape of probabilistic generative modeling that is theoretically simple and empirically effective.

\subsection{Limitations}\label{sec:limitations}

While \bsi{} applies directly to any continuous data in a fixed-dimensional space, we focused our experiments on low-resolution images up to 64\texttimes64 to compare the effects of the \bsi{} formulation against the baselines directly in data space.
Similar to other diffusion-like models, \bsi{} could be applied to high-resolution images by generating latent representations of an autoencoder \citep{rombach2022highresolution}.
Regarding model scale, our experiments cover the \unet{} of \citet{kingma2023variational} and the considerably larger \dit{}-L, while even larger backbones remain untested.
Further extension to non-image modalities and discrete data domains remains as future work.

\subsubsection*{Broader Impact Statement}

Generative modeling is widely used across many modalities today such as image and video data, audio signals or biological structures.
Therefore, advances in the methodology of generative modeling such as ours can have wide-ranging impacts on both desirable and undesirable developments, depending on the downstream application.
In particular, like other generative models of images, \bsi{} could be used to produce synthetic or manipulated media.
We believe that the most likely impact of our probabilistic perspective on generative modeling will be to inform other works in the field.

\subsubsection*{Acknowledgments}

This research was funded by the Bavarian State Ministry for Science and the Arts within the framework of the Geothermal Alliance Bavaria project.
We want to thank the Munich Center for Machine Learning for providing compute resources and are thankful to David L\"udke and Jonas Dornbusch for their valuable feedback.

\subsubsection*{Software}

For our results, we rely on excellent software packages, notably \texttt{numpy} \citep{harris2020array}, \texttt{pytorch} \citep{paszke2019pytorch}, \texttt{einops} \citep{rogozhnikov2022einops}, \texttt{matplotlib} \citep{hunter2007matplotlib}, \texttt{h5py} \citep{collette2013python}, \texttt{hydra} \citep{yadan2019hydra} and \texttt{jupyter} \citep{granger2021jupyter}.

\bibliography{main}
\bibliographystyle{tmlr}

\clearpage
\onecolumn
\appendix
\crefalias{section}{appendix}
\crefalias{subsection}{appendix}

\bookmarksetupnext{level=part}
\pdfbookmark{Appendix}{appendix}

\section{How BSI relates to \ldots}\label{sec:relations}

\subsection{Bayesian Flow Networks}\label{sec:bfns}

\bfns{} are a recent class of generative models for continuous and discrete data motivated from an information-theoretic perspective \citep{graves2023bayesian}.
In it, a sender communicates a data sample to a receiver while trying to minimize the transported data volume.
The sender compresses the data with entropy coding, so that minimizing the data volume is equivalent to the receiver maximizing the log-likelihood of that sample based on the information that it has received from the sender so far.
Finally, a sample can be generated when the receiver also assumes the role of the sender and repeatedly refines its belief.

\emph{For continuous data, \bfn{} sits at the deterministic end of the belief-prior range $[\lambda_{0}, \infty]$ of \cref{sec:prior-distribution}}.
To see this, we begin by choosing our belief prior $\pr(\vmu_{0})$ as $\Np(\vzero, \gamma_{0})$ and letting $\gamma_{0} \to \infty$, i.e.\ the initial belief mean will always be $\vmu_{0} = \vzero$.
With \cref{thm:elbo-encoder-closed-form}, this gives us
\begin{equation}
	\qr(\vmu_{\lambda} \mid \vx, \lambda) = \Np\bigg( \frac{\lambda - \lambda_{0}}{\lambda}\,\vx, \frac{\lambda^{2}}{\lambda - \lambda_{0}} \bigg).
\end{equation}
If we now define $\alpha = \lambda - \lambda_{0}$, choose the initial precision $\lambda_{0} = 1$ and write the Normal distribution in variance format, we see that
\begin{equation}
	\qr(\vmu_{\lambda} \mid \vx, \lambda) = \N\bigg( \frac{\alpha}{1 + \alpha} \vx, \frac{\alpha}{{(1 + \alpha)}^{2}} \bigg), \label{eq:bfn-flow-distribution}
\end{equation}
which equals the \bfn{} flow distribution $\pr_{\mathrm{F}}(\vtheta \mid \vx; t)$ \citep[Equation (76)]{graves2023bayesian} if we parametrize $\lambda$ (and therefore $\alpha$) in terms of $t \in [0, 1]$ as in \cref{sec:precision-encoding}.

Since a comprehensive description of \bfn{} would go beyond the scope of this work, we will only point out the correspondence between terms from \cref{sec:framework} and their \bfn{} counterparts without explaining them in detail.
For a complete description, we refer the reader to the original work \citep{graves2023bayesian}.

The current belief $(\vmu_{i}, \lambda_{i})$ is equivalent to the input distribution $\pr_{\mathrm{I}}$ \citep[Equation (43)]{graves2023bayesian}.
\cref{thm:posterior} is the equivalent of the Bayesian update function $h$ \citep[Section 4.2]{graves2023bayesian}.
A noisy measurement $\vy \sim \Np(\vx, \alpha)$ corresponds to the sender distribution $\pr_{\mathrm{S}}$ \citep[Equation (86)]{graves2023bayesian}, while a noisy measurement $\vy \sim \Np(\vxhat, \alpha)$ of the model's current prediction $\vxhat$ of the true sample corresponds to the receiver distribution $\pr_{\mathrm{R}}$ \citep[Equation (88)]{graves2023bayesian}.
The output distribution $\pr_{\mathrm{O}}$ and the Bayesian update distribution $\pr_{\mathrm{U}}$ are just intermediate terms to derive the model and appear neither in the final training nor sampling algorithm.

Fixing the initial belief to $\vmu_{0} = \vzero$ with infinite precision for \bfn{} recovers the behavior described by \citet[Figures 3 and 4]{graves2023bayesian} and shown in \cref{eq:bfn-flow-distribution} that the precision ${(1 + \alpha)}^{2} / \alpha$ of the flow / encoding distribution $\qr(\vmu_{\lambda} \mid \vx, \lambda)$ in the \elbo{} first falls and then rises again as $\alpha$ grows.
In contrast, with our belief prior $\pr(\vmu_{0}) = \Np(\vzero, \lambda_{0})$ of the same precision as the initial belief $(\vmu_{0}, \lambda_{0})$ as we choose it in \cref{sec:prior-distribution}, the precision of $\qr(\vmu_{\lambda} \mid \vx, \lambda)$ grows linearly in $\lambda$ (and $\alpha$) in lockstep with the precision of the belief $(\vmu_{i}, \lambda_{i})$.
We hypothesize that this makes learning for the model easier, because the noise level in its input varies linearly instead of non-linearly across noise levels.
Furthermore, in \bsi{}, the first sampling step will already contribute to drawing a random sample, since the initial input $\vmu_{0}$ to $\model$ is random.
In \bfn{}, the initial belief is fixed to $\vzero$, which makes the first sampling step deterministic and equal across all samples.

In our comparison to \dms{} in \cref{sec:diffusion-models}, we see that \bsi{} and \bfn{} also differ in their associated noising process.
There, we exploit that \bfn{} corresponds to $\gamma_{0} = \infty$ to derive its forward process and find that, while \bsi{}'s noising process, i.e.\ how one could go from a more precise belief back to a less precise one, does not form a Markov chain, \bfn{}'s does, making \bfn{} more similar to \dms{}.

\subsection{Diffusion Models}\label{sec:diffusion-models}

\dms{} are a widely used class of generative models built on the concept of inverting a diffusion process \citep{sohl-dickstein2015deep, ho2020denoising}.
Given a sample $\vx$, they define a Markov chain of increasingly noisy versions $\vx_{0}, \vx_{1}, \vx_{2}, \ldots$ of $\vx$ where $\vx_{0} = \vx$ and
\begin{equation}
	\pr(\vx_{i} \mid \vx_{i - 1}) = \N(a_{i}\vx_{i - 1}, b_{i}) \label{eq:diffusion-markov-chain}
\end{equation}
for some coefficients $a_{i}$ and $b_{i}$.
In training, a model learns to invert this Markov chain, which finally lets us generate data by sampling from a noise distribution and stepping along the learned, reverse Markov chain until we reach the data distribution.

While \dms{} initially achieved prominence in image generation \citep{dhariwal2021diffusion}, they have since been applied successfully across a variety of domains, such as text-to-image mapping \citep{saharia2022photorealistic}, fluid simulations \citep{lienen2024zero, saydemir2024unfolding}, adversarial attacks \citep{kollovieh2024assessing}, temporal \citep{ludke2023add} and general point processes \citep{ludke2024unlocking}, molecular dynamics \citep{lewis2025scalable}, molecular structure generation \citep{ayadi2024unified}, and time series forecasting \citep{kollovieh2023predict, kollovieh2024flow}.

\dms{} and \bsi{} are remarkably similar at first glance.
Both revolve around the concept of iteratively transforming noise into data samples, though \dms{} learn to invert a noising process, while \bsi{} uses posterior inference.
For training, both models aim to align a parametric distribution $\pr_{\vtheta}(\vx'' \mid \vx')$ with a distribution $\qr(\vx'' \mid \vx', \vx)$ that describes a less noisy version $\vx''$ of a noisy sample $\vx'$ given that the true sample is $\vx$.

However, conceptually, they approach sampling from two different perspectives.
\dms{} start with the so-called forward process, where signal is iteratively converted into noise forming a Markov chain of intermediate states as in \cref{eq:diffusion-markov-chain}.
Then, they revert this chain to derive the reverse process that enriches noise with data.
In contrast, \bsi{} defines the reverse process directly in the form of \cref{thm:update-marginal} and never uses the associated forward process directly.

We can revert \bsi{}'s process to derive its ``noising'' process.
This will let us see what \bsi{} would look like as a \dm{} and thus understand the relationship between the two.
Assume that our current belief is $(\vmu, \lambda = \lambda_{0} + \alpha)$ and we want to denoise further based on a sample $\vx$ and measurement precision $\alpha'$, i.e.\ update our belief to $(\vmu', \lambda' = \lambda_{0} + \alpha + \alpha')$.
The denoising process described by \cref{thm:update-marginal} tells us that
\begin{equation}
	\pr(\vmu' \mid \vmu, \vx) = \Np\big( \nicefrac{1}{\lambda'} \big[\lambda\vmu + \alpha'\vx\big], \nicefrac{{\lambda'}^{2}}{\alpha'} \big).
\end{equation}
To find the noising process, we revert this and get
\begin{equation}
	\pr(\vmu \mid \vmu', \vx) = \Np\Bigg( \xi\inv \bigg[ \frac{\lambda\lambda'}{\alpha'}\vmu' + \lambda\bigg( \frac{\alpha}{\alpha + \nicefrac{\lambda_{0}^{2}}{\gamma_{0}}} - 1 \bigg)\vx \bigg], \xi \Bigg) \label{eq:diffusion-bsi-forward}
\end{equation}
where $\xi = \lambda^{2}\big( {(\alpha + \nicefrac{\lambda_{0}^{2}}{\gamma_{0}})}\inv + {\alpha'}\inv \big)$ and $\gamma_{0}$ is the precision of the initial belief prior $\pr(\vmu_{0}) = \Np(\vzero, \gamma_{0})$.
Find the proof at the end of this section.

Plugging in $\gamma_{0} = \lambda_{0}$, we get that the noising process of \bsi{} is
\begin{equation}
	\pr(\vmu \mid \vmu', \vx) = \Np\Bigg( \xi\inv \bigg[ \frac{\lambda\lambda'}{\alpha'}\vmu' - \lambda_{0}\vx \bigg], \xi \Bigg) \quad \text{where} \quad \xi = \lambda\bigg(1 + \frac{\lambda}{\alpha'}\bigg).
\end{equation}
Note that this distribution depends on $\vx$ since $\lambda_{0} > 0$.
Therefore, \bsi{}'s forward process would not be Markov, i.e.\ we cannot add more noise to a belief state without knowing the sample $\vx$ that the belief state originated from.
While \dms{} with non-Markov forward processes exist \citep{song2021denoising, chen2024fast}, they are uncommon.
In conclusion, we see that \bsi{} can be represented as a \dm{}, though with a rather complex, non-Markovian forward process.

As we have shown in \cref{sec:bfns}, \bfns{} sit at the $\gamma_{0} = \infty$ end of the belief-prior range in \cref{sec:prior-distribution}.
Curiously, \cref{eq:diffusion-bsi-forward} shows that, among the isotropic Normal priors of \cref{sec:prior-distribution}, this is the only one for which the associated forward process is Markov, because the coefficient of $\vx$ becomes $0$.
This agrees with \citet{xue2024unifying}, who have shown that \bfns{} admit a formulation based on \sdes{}, like score-based \dms{} do.

\begin{proof}[Proof of \cref{eq:diffusion-bsi-forward}]
	We know from \cref{thm:elbo-encoder-closed-form} that
	\begin{equation}
		\qr(\vmu \mid \vx, \lambda) = \Np\bigg( \frac{\lambda - \lambda_{0}}{\lambda}\,\vx, \frac{\lambda^{2}}{\lambda - \lambda_{0} + \nicefrac{\lambda_{0}^{2}}{\gamma_{0}}} \bigg) = \Np\bigg( \frac{\alpha}{\lambda}\,\vx, \frac{\lambda^{2}}{\alpha + \nicefrac{\lambda_{0}^{2}}{\gamma_{0}}} \bigg)
	\end{equation}
	and from \cref{thm:update-marginal} that
	\begin{equation}
		\pr(\vmu' \mid \vmu, \vx) = \Np\big( \nicefrac{1}{\lambda'} \big[\lambda\vmu + \alpha'\vx\big], \nicefrac{{\lambda'}^{2}}{\alpha'} \big).
	\end{equation}
	Therefore, $\pr(\vmu, \vmu' \mid \vx)$ is a Gaussian linear system and we can use \citep[Equation (4.125)]{murphy2012machine} to see that
	\begin{equation}
		\pr(\vmu \mid \vmu', \vx) = \Np(\vnu, \xi)
	\end{equation}
	with
	\begin{equation}
		\xi = \lambda^{2}{\bigg(\alpha + \frac{\lambda_{0}^{2}}{\gamma_{0}}\bigg)}\inv + {\bigg(\frac{\lambda}{\lambda'}\bigg)}^{2}\frac{{\lambda'}^{2}}{\alpha'} = \lambda^{2}\Big( {(\alpha + \nicefrac{\lambda_{0}^{2}}{\gamma_{0}})}\inv + {\alpha'}\inv \Big)
	\end{equation}
	and
	\begin{align}
		\vnu & = \xi\inv \Bigg[ \frac{\lambda}{\lambda'}\frac{{\lambda'}^{2}}{\alpha'} \bigg( \vmu' - \frac{\alpha'}{\lambda'}\vx \bigg) + \lambda^{2}{(\alpha + \nicefrac{\lambda_{0}^{2}}{\gamma_{0}})}\inv\,\nicefrac{\alpha}{\lambda}\,\vx \Bigg] \\
		     & = \xi\inv \Bigg[ \frac{\lambda\lambda'}{\alpha'}\vmu' + \lambda\bigg( \frac{\alpha}{\alpha + \nicefrac{\lambda_{0}^{2}}{\gamma_{0}}} - 1 \bigg)\vx \Bigg].
	\end{align}
\end{proof}

\subsection{Stochastic Interpolants}\label{sec:stochastic-interpolants}

Stochastic interpolants are a broad class of continuous-time stochastic processes that can interpolate between any two probability distributions $\rho_{0}$ and $\rho_{1}$ \citep{albergo2025stochastic}.
They also prescribe how to learn the interpolants' dynamics to construct generative models and it is instructive to see how they relate to \bsi{}.
The subclass of \emph{spatially linear one-sided interpolants} assumes that $\rho_{0}$ is a standard Normal distribution and defines the interpolant
\begin{equation}
	\vx_{t} = \alpha(t)\vz_{t} + \beta(t)\vx_{1} \label{eq:stochastic-interpolant}
\end{equation}
where $\vx_{1} \sim \rho_{1}$ and $\vz_{t} \sim \N(\vzero, \mI)$.
Given that $\alpha$ and $\beta$ are smooth, non-negative functions with $\alpha(0) = \beta(1) = 1$ and $\alpha(1) = \beta(0) = 0$, $\vx_{t}$ smoothly interpolates between a standard Normal and the data distribution $\rho_{1}$.

For \bsi{}, we can interpret the belief mean $\vmu_{\lambda}$ for a data sample $\vx$
\begin{equation}
	\vmu_{\lambda} = \frac{1}{\sqrt{\lambda}}\vz + \frac{\lambda - \lambda_{0}}{\lambda}\vx
\end{equation}
as an interpolant that is normally distributed with mean $\vzero$ and precision $\lambda_{0}$ at $\lambda = \lambda_{0}$ and equals the data sample $\vx$ at $\lambda = \infty$.
We could rewrite this as an interpolant in the above sense on $[0, 1]$ by parameterizing $\lambda$ as a strictly increasing function $\lambda(t) : [0, 1] \to \Rplus$ with $\lambda(0) = \lambda_{0}$ and $\lim_{t \to 1} \lambda(t) = \infty$, e.g.\ $\lambda(t) = \lambda_{0} - \log(1 - t)$, similar to the mapping between score-based diffusion and stochastic interpolants \citep[Section 5.1]{albergo2025stochastic}.
But to avoid the scaling and correction factors, we will consider it an interpolant on $[\lambda_{0}, \infty]$ instead with $\alpha(\lambda) = 1 / \sqrt{\lambda}$ and $\beta(\lambda) = (\lambda - \lambda_{0}) / \lambda$.
We will furthermore write $\alpha(\lambda)$ and $\beta(\lambda)$ as $\alpha$ and $\beta$ to reduce visual clutter.

\citet[Section 4.4]{albergo2025stochastic} show that the probability path $\rho(\lambda, \vmu_{\lambda})$ of the interpolant solves the transport equation $\partial_{\lambda}\rho + \nabla \cdot (b\rho) = 0$ with the velocity field
\begin{equation}
	b(\lambda, \vmu) = \frac{\dot{\alpha}}{\alpha}\vmu + \bigg(\dot{\beta} - \frac{\dot{\alpha}}{\alpha}\beta\bigg)\eta(\lambda, \vmu)
\end{equation}
and its score $\nabla \log\rho(\lambda, \vmu)$ is given by
\begin{equation}
	s(\lambda, \vmu) = -\frac{\vmu - \beta\eta(\lambda, \vmu)}{\alpha^{2}}.
\end{equation}
$\eta(\lambda, \vmu) = \E[\vx \mid \vmu_{\lambda} = \vmu]$ is the denoiser, i.e.\ the expected data sample that led to belief $\vmu$ at precision $\lambda$.
Note that $\eta(\lambda, \vmu)$ is learned with a model $\hat{\eta}$, which is equivalent to $\model(\vmu, \lambda)$ in \bsi{} and fit by minimizing \citep[Eq. (4.21)]{albergo2025stochastic}
\begin{equation}
	\mathcal{L}(\hat{\eta}) = \int_{\lambda_{0}}^{\lambdaM} \E\bigg[\frac{1}{2}{\|\hat{\eta}(\lambda, \vmu)\|}_{2}^{2} - \vx \cdot \hat{\eta}(\lambda, \vmu)\bigg]\,\mathrm{d}\lambda.
\end{equation}
$\mathcal{L}(\hat{\eta})$ is equivalent to $\lmeasureinf$ in \cref{thm:infinite-elbo} up to a constant factor and offset.

With the velocity field and score, we can write down the forward \sde{} corresponding to the probability path $\rho$ as \citep[Corollary 18]{albergo2025stochastic}
\begin{equation}
	\mathrm{d}\vmu_{\lambda} = b_{\mathrm{F}}(\lambda, \vmu_{\lambda})\,\mathrm{d}\lambda + \sqrt{2\varepsilon(\lambda)}\,\mathrm{d}W_{\lambda}
\end{equation}
where $W_{\lambda}$ is Brownian motion, $\varepsilon(\lambda) : \R \to \Rplus$ is any noise level specification and
\begin{equation}
	\begin{aligned}
		b_{\mathrm{F}}(\lambda, \vmu) & = b(\lambda, \vmu) + \varepsilon(\lambda)s(\lambda, \vmu)                                                                                                                                                     \\
		                              & = \bigg(\frac{\dot{\alpha}}{\alpha} - \frac{\varepsilon}{\alpha^{2}}\bigg)\vmu + \bigg(\frac{\varepsilon}{\alpha^{2}} - \frac{\dot{\alpha}}{\alpha} + \frac{\dot{\beta}}{\beta}\bigg)\beta\eta(\lambda, \vmu)
	\end{aligned}
\end{equation}
is the forward drift.
If we plug in $\alpha$ and $\beta$, we get
\begin{equation}
	\mathrm{d}\vmu_{\lambda} = \bigg[-\bigg(\frac{1}{2\lambda} + \varepsilon\lambda\bigg)\vmu_{\lambda} + \bigg(\varepsilon\lambda + \frac{1}{2\lambda} + \frac{\lambda_{0}}{\lambda(\lambda - \lambda_{0})}\bigg)\frac{\lambda - \lambda_{0}}{\lambda}\eta(\lambda, \vmu_{\lambda})\bigg]\,\mathrm{d}\lambda + \sqrt{2\varepsilon}\,\mathrm{d}W_{\lambda}.
\end{equation}
Since this holds for any non-negative $\varepsilon$, we can choose $\varepsilon = \frac{1}{2\lambda^{2}}$ to simplify the equation to
\begin{equation}
	\mathrm{d}\vmu_{\lambda} = \frac{1}{\lambda}\big[\eta(\lambda, \vmu_{\lambda}) - \vmu_{\lambda}\big]\,\mathrm{d}\lambda + \frac{1}{\lambda} \mathrm{d}W_{\lambda}. \label{eq:si-forward}
\end{equation}

We can now sample from the learned stochastic interpolant by integrating \cref{eq:si-forward} from $\lambda_{0}$ to $\lambda_{M}$ \citep[Algorithm 5]{albergo2025stochastic}.
Let's say we are at precision $\lambda$ with state $\vmu_{\lambda}$ and want to move ahead by a step of length $\alpha$.
With the Euler-Maruyama method suggested by \citet{albergo2025stochastic}, the integration step becomes
\begin{equation}
	\begin{aligned}
		\vmu_{\lambda + \alpha} & = \vmu_{\lambda} + \alpha \cdot \frac{1}{\lambda}\big[\hat{\eta}(\lambda, \vmu_{\lambda}) - \vmu_{\lambda}\big] + \frac{\sqrt{\alpha}}{\lambda}\,\vepsilon \\
		                        & = \frac{\lambda - \alpha}{\lambda}\vmu_{\lambda} + \frac{\alpha}{\lambda}\hat{\eta}(\lambda, \vmu_{\lambda}) + \frac{\sqrt{\alpha}}{\lambda}\,\vepsilon
	\end{aligned}\label{eq:si-step}
\end{equation}
where $\vepsilon \sim \N(\vzero, \mI)$.
This is almost the same as the \bsi{} sampling step
\begin{equation}
	\begin{aligned}
		\vmu_{i} & = \frac{\lambda_{i-1} \vmu_{i-1} + \alpha_{i}\big(\vxhat_{i-1} + \sqrt{\frac{1}{\alpha_{i}}}\vepsilon_{i}\big)}{\lambda_{i - 1} + \alpha_{i}}               \\
		         & = \frac{\lambda_{i} - \alpha_{i}}{\lambda_{i}} \vmu_{i-1} + \frac{\alpha_{i}}{\lambda_{i}}\vxhat_{i-1} + \frac{\sqrt{\alpha_{i}}}{\lambda_{i}}\vepsilon_{i}
	\end{aligned}\label{eq:si-bsi-step}
\end{equation}
in \cref{alg:generate-specific}.
The subtle difference is that \cref{eq:si-step} uses the current $\lambda$ to compute $\vmu_{\lambda + \alpha}$ whereas \cref{eq:si-bsi-step} uses the next $\lambda_{i} = \lambda_{i - 1} + \alpha_{i}$ for $\vmu_{i}$.
This difference reflects that \bsi{} employs an exact Bayesian update rather than a first-order Euler-Maruyama approximation.

In summary, we can write \bsi{} as a spatially linear one-sided stochastic interpolant, though two important differences remain.
First, while the sampling steps \cref{eq:si-step,eq:si-bsi-step} are equivalent in the continuous limit of $\alpha \to 0$, they differ in practice due to their derivation from an \sde{} discretization and posterior inference, respectively.
Second, stochastic interpolants require the interpolation's endpoints to equal the noise and data distribution exactly, so that the data end corresponds to $\lambda = \infty$ in the above formulation.
In contrast, \bsi{} only infers the sample $\vx$ up to a maximum precision of $\lambdaM$, e.g.\ precisely enough to identify the exact color in an image with 8-bit color channels.

\subsection{Posterior Mean Matching}\label{sec:pmm}

\citet{salazar2025posterior} have introduced \pmm{}.
Like \bsi{}, they formulate generative modeling as a posterior inference problem where the sample is inferred from a sequence of noisy observations.
In contrast to \bsi{}, they begin sample generation from a fixed initial belief $\vmu_{0} = \vzero$ instead of sampling the initial belief itself from a hyper-prior $\pr(\vmu_{0})$.
This is the same choice as \bfn{} and, in fact, the Normal-Normal \pmm{} model for continuous data equals the deterministic end of the belief-prior range that we derive in \cref{sec:prior-distribution}.
The elimination of the hyper-prior lets \citeauthor{salazar2025posterior} extend \pmm{} to conjugate belief state and observation distribution pairs other than Normal-Normal, such as Dirichlet-Categorical.
Since the construction in \cref{sec:framework} permits a Dirac delta hyper-prior, i.e.\ $\gamma_{0} \to \infty$, the connection to \pmm{} might offer a way to generalize \bsi{} to discrete data.

\subsection{Sequential Monte Carlo}\label{sec:smc}

If the target distribution is specified as an unnormalized density $\tilde{\pr}$ instead of through a dataset of samples, we can sample from $\tilde{\pr}$ with \smc{} methods.
Similar to \bsi{}, such methods also often rely on importance sampling for variance reduction.
However, in high dimensions, the variance of an importance sampling estimate can increase exponentially with the dimension.
\citet{chen2025sequential} propose to solve this problem with a resampling scheme combined with stochastic diffusion.
\citet{wu2025reverse} use reverse diffusion dynamics as the proposal of an \smc{} sampler and correct the resulting bias through importance weighting and resampling against a sequence of intermediate targets.
Note that in \bsi{} we apply importance sampling to estimate the \elbo{} over the space of precision levels $\lambda$, a scalar, and thus do not experience exploding variance.

\section{ELBO in Bits per Dimension}\label{sec:bits-per-dimension}

A common metric in probabilistic modeling is the negative log-likelihood of unseen data.
The benefits of this metric are that it is theoretically motivated by the probabilistic framework and it can be computed across domains regardless of data modality.
If the negative log-likelihood is small, the generative model assigns high likelihood to the unseen data and can thus be regarded as a good model (though likelihood and sample quality are not necessarily the same thing \citep{theis2016note}).
For models that come with an \elbo{} like \bsi{}, we can use it to upper bound the negative log-likelihood to compare against other \elbo{}-based or exact-likelihood models.

The negative log-likelihood is usually reported in bits per pixel, per color channel or, in general, per dimension.
This unit comes from the fact that, for a finite symbol alphabet $\sS$, an entropy coder could use the model to encode samples $\vx \in \sS^{\ndim}$ from the data distribution asymptotically in $-\negthinspace\log_{2} \pr_{\vtheta}(\vx) / \ndim$ bits per dimension \citep{duda2015use}.
Note that the underlying space $\sS$ must be discrete.
If it were continuous, $\pr_{\vtheta}(\vx)$ would be a density and the theory would predict that we could compress $\vx$ into a negative number of bits.

The discreteness requirement is a natural fit for many domains.
While, for example, images are usually treated as tensors with continuous color values, the colors are actually stored as discrete values in the range $[0, 2^{8} - 1]$ for 8-bit images.
Similarly, audio data is a sequence of discrete values in, for example, a 16-bit range.

Let's say that $\sS$ is the set of integers $\{0, \ldots, \nbuckets - 1\}$.
Then we can compute an upper bound on the bits needed to encode $\vx \in \sS^{\ndim}$ by
\begin{equation}
	-\log_{2}\pr(\vx) \le \frac{1}{\log(2)} \big(\lrecon' + \lmeasureinf\big)
\end{equation}
as per \cref{thm:elbo,thm:infinite-elbo}.
The division by $\log(2)$ converts the logarithms in $\lrecon'$ and $\lmeasureinf$ to base 2.
$\lrecon'$ is the same as $\lrecon$ but with a discretized Normal likelihood to account for the discrete nature of $\vx$, i.e.
\begin{equation}
	\lrecon' = \E_{\qr(\vmu_{\lambdaM} \mid \vx, \lambdaM)}\big[{\shortminus}\negthinspace\log\Np'(\vx \mid \vxhat_{\lambdaM}, \alphaR)\big]
\end{equation}
where
\begin{equation}
	\Np'(\evx_{j} \mid \vxhat_{\lambdaM}, \alphaR) = \varPhi(r_{j} \mid \vxhat_{\lambdaM}, \alphaR) - \varPhi(l_{j} \mid \vxhat_{\lambdaM}, \alphaR).
\end{equation}
$\varPhi(r_{j} \mid \vxhat_{\lambdaM}, \alphaR)$ is the \cdf{} of $\Np(\vxhat_{\lambdaM}, \alphaR)$ and $l_{j}$ and $r_{j}$ are the boundaries of the discretization interval containing $\evx_{j}$, i.e.
\begin{equation}
	l_{j} = \begin{cases}
		-\infty                                              & \text{if}\ \evx_{j} < \frac{1}{2}               \\
		\nbuckets - \frac{3}{2}                              & \text{if}\ \evx_{j} \ge \nbuckets - \frac{3}{2} \\
		\lfloor \evx_{j} - \frac{1}{2} \rfloor + \frac{1}{2} & \text{otherwise}
	\end{cases}
	\quad\text{and}\quad
	r_{j} = \begin{cases}
		\infty                                               & \text{if}\ \evx_{j} \ge \nbuckets - \frac{3}{2} \\
		\frac{1}{2}                                          & \text{if}\ \evx_{j} < \frac{1}{2}               \\
		\lfloor \evx_{j} + \frac{1}{2} \rfloor + \frac{1}{2} & \text{otherwise.}
	\end{cases}
\end{equation}

$\lmeasureinf$ is usually not discretized during \elbo{} computation as the latent variables only enter as a mean squared error instead of a log-likelihood.
In a practical implementation, the latent variable distributions would need to be discretized as well, decreasing the \elbo{} slightly \citep{kingma2023variational, townsend2019hilloc}.
If $\vx$ is discretized to a different set of discrete symbols, e.g.\ numbers between $-1$ and $1$ instead of the integers $\sS$, the boundaries of the discretization intervals and bin widths in the discretized Normal distribution have to be adapted accordingly.

\section{Preconditioning Derivation}\label{sec:preconditioning-derivation}

We will assume in this section that the data is normalized such that $\E[\vx] = \vzero$ and $\var[\vx] = \mI$.

Assume that we have a current belief $(\vmu, \lambda)$.
We derive the parameters $\preskip$, $\preout$ and $\prein$ of the preconditioned model
\begin{equation}
	\model(\vmu, \lambda) = \preskip \vmu + \preout \model'(\prein\vmu, \lambda) \label{eq:precond-model}
\end{equation}
analogously to \citet{karras2022elucidating}.
However, while we proceed in the same way, the resulting parameters for \bsi{} differ from \citet{karras2022elucidating} because \bsi{} is not included in the family of \dms{} that \citet{karras2022elucidating} consider, see \cref{sec:diffusion-models}.

First, we require that $\var_{\vx}[\prein\vmu] = \mI$ for all $\lambda$.
We know from \cref{thm:bsi-elbo-encoder-closed-form} that
\begin{equation}
	\qr(\vmu \mid \vx, \lambda) = \Np\big(\nicefrac{(\lambda - \lambda_{0})}{\lambda}\,\vx, \lambda\big).
\end{equation}
Therefore, $\pr(\vx, \vmu)$ is a Gaussian linear system and \citep[Equation (4.126)]{murphy2012machine} tells us that the variance of the marginal distribution of $\vmu$ is
\begin{equation}
	\var_{\vx}[\vmu] = \bigg(\lambda\inv + \frac{{(\lambda - \lambda_{0})}^{2}}{\lambda^{2}}\bigg)\mI.
\end{equation}
By plugging this into our requirement
\begin{equation}
	\var_{\vx}[\prein\vmu] = \prein^{2}\var_{\vx}[\vmu] = \mI,
\end{equation}
we get immediately that
\begin{equation}
	\prein = {\bigg(\lambda\inv + \frac{{(\lambda - \lambda_{0})}^{2}}{\lambda^{2}}\bigg)}\bigprsqrt = {\bigg(\underbrace{1 + \frac{{(\lambda - \lambda_{0})}^{2}}{\lambda}}_{\eqcolon \kappa}\bigg)}\bigprsqrt \lambda\powsqrt = \sqrt{\nicefrac{\lambda}{\kappa}}.
\end{equation}

Next, we want the actual prediction target of $\model'$ during training to have unit variance, too.
In training, we optimize the \elbo{} from \cref{thm:infinite-elbo}, which comes down to minimizing
\begin{equation}
	\big\|\vx - \model(\vmu, \lambda)\big\|_{2}^{2}
\end{equation}
up to constant factors only depending on $\lambda$.
If we plug in \cref{eq:precond-model} and isolate $\model'$, this distance becomes
\begin{equation}
	\big\|\vx - \preskip \vmu - \preout \model'(\prein\vmu, \lambda)\big\|_{2}^{2} = \preout^{2}\big\|\model'(\prein\vmu, \lambda) - \preout\inv(\vx - \preskip \vmu)\big\|_{2}^{2}.
\end{equation}
From this, we identify $\preout\inv(\vx - \preskip \vmu)$ as the actual training target for $\model'$.
For the rest of this derivation, we use the shorthand $\alpha = \lambda - \lambda_{0}$ for the measurement precision accumulated in our belief $(\vmu, \lambda)$.
After \cref{thm:bsi-elbo-encoder-closed-form}, we can write $\vmu$ as $\nicefrac{\alpha}{\lambda}\,\vx + \vz$ where $\vz \sim \Np(\vzero, \lambda)$ and find that the variance of the training target is
\begin{align}
	\begin{aligned}
		\var_{\vx, \vz}[\preout\inv(\vx - \preskip \vmu)] & = \preout^{\shortminus{}2}\var_{\vx, \vz}\bigg[\vx - \preskip \bigg(\frac{\alpha}{\lambda}\,\vx + \vz\bigg)\bigg]              \\
		                                                  & = \preout^{\shortminus{}2}\var_{\vx, \vz}\bigg[\bigg( 1 - \preskip \frac{\alpha}{\lambda} \bigg)\vx - \preskip\vz\bigg]        \\
		                                                  & = \preout^{\shortminus{}2} \bigg[ {\bigg( 1 - \preskip \frac{\alpha}{\lambda} \bigg)}^{2} + \preskip^{2}\lambda\inv \bigg] \mI.
	\end{aligned}
\end{align}
If we now require the effective training target to have unit variance, we see that
\begin{equation}
	\preout^{2} = {\bigg( 1 - \preskip \frac{\alpha}{\lambda} \bigg)}^{2} + \preskip^{2}\lambda\inv = \bigg[1 + \frac{\alpha^{2}}{\lambda}\bigg]\frac{1}{\lambda}\preskip^{2} - 2\frac{\alpha}{\lambda}\preskip + 1. \label{eq:precond-precout-general}
\end{equation}

Following \citet{karras2022elucidating}, we now choose $\preskip$ to minimize the impact of errors in the output of $\model'$ by minimizing the magnitude of $\preout$.
$\preout^{2}$ is a polynomial in $\preskip$ with positive leading coefficient, so we can find the minimizer as the root of
\begin{equation}
	\frac{1}{2}\,\frac{\mathrm{d}\preout^{2}}{\mathrm{d}\preskip} = \bigg[1 + \frac{\alpha^{2}}{\lambda}\bigg]\frac{1}{\lambda}\preskip - \frac{\alpha}{\lambda},
\end{equation}
which is at
\begin{equation}
	\preskip = {\bigg[1 + \frac{\alpha^{2}}{\lambda}\bigg]}\inv \alpha = \kappa\inv\alpha = \frac{\alpha}{\kappa}.
\end{equation}

Finally, we can plug $\preskip$ into \cref{eq:precond-precout-general} to get
\begin{equation}
	\begin{aligned}
		\preout^{2} & = \kappa\kappa^{\shortminus{}2} \frac{\alpha^{2}}{\lambda} - 2\frac{\alpha}{\lambda}\kappa\inv \alpha + 1 = \kappa\inv \bigg( \frac{\alpha^{2}}{\lambda} - 2\frac{\alpha^{2}}{\lambda} + \bigg[1 + \frac{\alpha^{2}}{\lambda}\bigg] \bigg) = \kappa\inv
	\end{aligned}
\end{equation}
and consequently $\preout = \kappa\rsqrt = \sqrt{\nicefrac{1}{\kappa}}$.

\section{Proofs}\label{sec:proofs}

\subsection{\texorpdfstring{Proof of \cref{thm:elbo}}{Proof of Theorem \ref{thm:elbo}}}\label{sec:proof-of-elbo}

We will begin with some auxiliary insights.
First, we consider the marginal distribution of the updated belief $(\vmu', \lambda')$.
This means that our current belief about a sample $\vx$ is $(\vmu, \lambda)$ and now we want to know the distribution of $\vmu'$ after updating $\vmu$ with \cref{thm:posterior} marginalized over all possible noisy measurements $\vy$ with precision $\alpha$.
Note that $\lambda'$ is deterministic as it neither depends on $\vx$ nor $\vy$.

\begin{lemma}[Update Marginal]\label{thm:update-marginal}
	Let $\vx, \vmu \in \R^{\ndim}$ and $\lambda, \alpha \in \Rplus$.
	Then the distribution of the posterior belief mean $\vmu'$ marginalized over all measurements $\vy$ made with precision $\alpha$ is
	\begin{equation}
		\pr(\vmu' \mid \vmu, \vx, \alpha) = \E_{\vy \sim \Np(\vx, \alpha\mI)}\big[ \pr(\vmu' \mid \vmu, \vx, \alpha, \vy) \big] = \Np\big( \nicefrac{1}{\lambda'} \big[\lambda\vmu + \alpha\vx\big], \nicefrac{{\lambda'}^{2}}{\alpha} \big). \label{eq:update-marginal}
	\end{equation}
\end{lemma}
\begin{proof}
	The noisy measurement is a Normal random variable $\vy \sim \Np(\vx, \alpha)$ and the mean of our posterior belief $(\vmu', \lambda')$ after observing $\vy$ is the deterministic linear transformation
	\begin{equation}
		\vmu' = \nicefrac{1}{\lambda'} \left[ \lambda\vmu + \alpha\vy \right]
	\end{equation}
	of this random variable.
	The statement follows immediately by the linear transformation property of the Normal distribution.
\end{proof}

From this, we can see that the update marginal from multiple intermediate measurements is the same as from a single measurement with the combined precision of the intermediate measurements.

\begin{lemma}\label{thm:multi-update-marginal}
	Let $\vx, \vmu, \vmu', \vmu'' \in \R^{\ndim}$ and $\lambda, \alpha, \alpha' \in \Rplus$.
	$\vmu'$ is the posterior belief mean after a measurement with precision~$\alpha$ and $\vmu''$ the posterior belief mean after a second, subsequent measurement with precision~$\alpha'$.
	Then we have that the marginal distribution of the second update is
	\begin{equation}
		\E_{\pr(\vmu' \mid \vmu, \vx, \alpha)}[\pr(\vmu'' \mid \vmu', \vx, \alpha')] = \pr(\vmu'' \mid \vmu, \vx, \alpha + \alpha').
	\end{equation}
\end{lemma}
\begin{proof}
	We know from \cref{thm:update-marginal} that $\vmu'$ is a random variable
	\begin{equation}
		\pr(\vmu' \mid \vmu, \vx, \alpha) = \Np\big( \underbrace{\nicefrac{1}{\lambda'} \big[\lambda\vmu + \alpha\vx\big]}_{\eqcolon \vnu}, \underbrace{\nicefrac{{\lambda'}^{2}}{\alpha}}_{\eqcolon \xi} \big)
	\end{equation}
	and $\vmu''$ is a random variable that depends linearly on $\vmu'$
	\begin{equation}
		\pr(\vmu'' \mid \vmu', \vx, \alpha') = \Np\big( \nicefrac{1}{\lambda''} \big[\lambda'\vmu' + \alpha'\vx\big], \nicefrac{{\lambda''}^{2}}{\alpha'} \big).
	\end{equation}
	As such, they jointly form a Gaussian linear system for which the marginal distribution of $\vmu''$ is \citep[Equation (4.126)]{murphy2012machine}
	\begin{equation}
		\E_{\pr(\vmu' \mid \vmu, \vx, \alpha)} [\pr(\vmu'' \mid \vmu', \vx, \alpha')] = \N\bigg( \nicefrac{1}{\lambda''} \big[\lambda'\vnu + \alpha'\vx\big], \frac{\alpha'}{{\lambda''}^{2}} + \frac{{\lambda'}^{2}}{{\lambda''}^{2}\xi} \bigg). \label{eq:multi-update-marginal-marginal}
	\end{equation}

	Plugging $\vnu$ into the mean expression and simplifying yields the marginal mean
	\begin{equation}
		\nicefrac{1}{\lambda''} \big[\lambda\vmu + (\alpha + \alpha')\vx\big].
	\end{equation}
	Similarly, plugging $\xi$ into the covariance expression and simplifying yields the marginal covariance
	\begin{equation}
		\frac{\alpha + \alpha'}{{\lambda''}^{2}}.
	\end{equation}
	If we now recall from \cref{thm:posterior} that
	\begin{equation}
		\lambda' = \lambda + \alpha \quad \text{and} \quad \lambda'' = \lambda' + \alpha' = \lambda + \alpha + \alpha',
	\end{equation}
	we can identify \cref{eq:multi-update-marginal-marginal} as $\pr(\vmu'' \mid \vmu, \vx, \alpha + \alpha')$.
\end{proof}

This trivially generalizes to any finite sequence of measurements, which can be collapsed into a single measurement with the total precision instead.

We will furthermore need to know the KL divergence between the update marginal distributions of the same belief but based on two different samples $\vx$ and $\vx'$.

\begin{lemma}\label{thm:kl-div-marginal-update}
	Let $\vx, \vx', \vmu \in \R^{\ndim}$ and $\lambda, \alpha \in \Rplus$.
	Then
	\begin{equation}
		\kl(\pr(\vmu' \mid \vmu, \vx, \alpha), \pr(\vmu' \mid \vmu, \vx', \alpha)) = \nicefrac{1}{2}\,\alpha \|\vx - \vx'\|_{2}^{2}.
	\end{equation}
\end{lemma}
\begin{proof}
	Both update marginal distributions -- with $\vx$ and $\vx'$ -- are Normal distributions of equal precision $\xi \coloneq \frac{{\lambda'}^{2}}{\alpha}$ as given by \cref{thm:update-marginal} and respective means of
	\begin{equation}
		\vnu = \nicefrac{1}{\lambda'} \big[\lambda\vmu + \alpha\vx\big] \quad \text{and} \quad \vnu' = \nicefrac{1}{\lambda'} \big[\lambda\vmu + \alpha\vx'\big].
	\end{equation}
	As a consequence, the closed form solution for the KL divergence between two equal-covariance Normal distributions becomes
	\begin{equation}
		\begin{aligned}
			\kl(\pr(\vmu' \mid \vmu, \vx, \alpha), \pr(\vmu' \mid \vmu, \vx', \alpha)) & = \frac{1}{2} (\vnu - \vnu')\T \xi (\vnu - \vnu')                                         \\
			                                                                           & = \frac{1}{2} (\vx - \vx')\T \alpha {\lambda'}\inv \xi {\lambda'}\inv \alpha (\vx - \vx') \\
			                                                                           & = \frac{1}{2} (\vx - \vx')\T \alpha (\vx - \vx')                                          \\
			                                                                           & = \frac{1}{2} \alpha \|\vx - \vx'\|_{2}^{2}.
		\end{aligned}
	\end{equation}
\end{proof}

Equipped with these, we can derive the \elbo{}.

\thmelbo*
\begin{proof}
	For any distribution $\pr(\vx)$ and any latent variable $\vz$, i.e.\ any choice of prior $\pr(\vz)$, encoding distribution $\qr(\vz \mid \vx)$ and likelihood $\pr(\vx \mid \vz)$, we have the variational lower bound
	\begin{equation}
		\log\pr(\vx) \ge -\E_{\qr(\vz \mid \vx)} [-\log \pr(\vx \mid \vz)] - \kl(\qr(\vz \mid \vx), \pr(\vz)) \label{eq:elbo-proof-elbo}
	\end{equation}
	on $\log\pr(\vx)$ \citep{kingma2013autoencoding}.
	In particular, we can choose our sequence of beliefs as the latent variable $\vz = \{\vmu_{0}, \ldots, \vmu_{\nrounds}\}$ and define the likelihood of $\vx$ under this latent variable as
	\begin{equation}
		\pr(\vx \mid \vz) = \Np(\vx \mid \vxhat_{\nrounds}, \alphaR). \label{eq:elbo-proof-decoder}
	\end{equation}
	Remember that $\vxhat_{k} = \model(\vmu_{k}, \lambda_{k})$ is the model's estimate of $\vx$.

	Since the belief means $\vmu_{1}, \ldots, \vmu_{k}$ are updated only based on their predecessor after \cref{thm:posterior}, they form a Markov chain conditional on $\vx$ and we can write the encoding distribution as
	\begin{equation}
		\qr(\vz \mid \vx) = \pr(\vmu_{0})\, \prod_{i = 1}^{\nrounds} \pr(\vmu_{i} \mid \vmu_{i - 1}, \vx, \alpha_{i}). \label{eq:elbo-proof-encoding}
	\end{equation}
	Each $\pr(\vmu_{i} \mid \vmu_{i - 1}, \vx, \alpha_{i})$ is the update marginal of $\vmu_{i - 1}$ over all possible noisy measurements of $\vx$ with precision $\alpha_{i}$ from \cref{thm:update-marginal}.
	Our encoding distribution is ignorant about the influence of $\vx$ on the initial belief $\vmu_{0}$, because there is no closed form for $\pr(\vmu_{0} \mid \vx)$.
	Since we can choose any encoding, not encoding $\vx$ in $\vmu_{0}$ at all is valid.

	If we now plug \cref{eq:elbo-proof-encoding} into the first term of \cref{eq:elbo-proof-elbo}, we get
	\begin{equation}
		\E_{\qr(\vz \mid \vx)} [{\shortminus}\negthinspace\log \pr(\vx \mid \vz)] = \E_{\pr(\vmu_{0})} \E_{\pr(\vmu_{1} \mid \vmu_{0}, \vx, \alpha_{1})}\,\ldots\,\E_{\pr(\vmu_{\nrounds} \mid \vmu_{\nrounds - 1}, \vx, \alpha_{\nrounds})} [{\shortminus}\negthinspace\log \pr(\vx \mid \vz)].
	\end{equation}
	The intermediate expectations collapse into a single measurement with the sum of all precisions $\bar{\alpha}_{i} = \sum_{j = 1}^{i} \alpha_{j}$ according to \cref{thm:multi-update-marginal}, because $\vmu_{1}, \ldots, \vmu_{\nrounds - 1}$ do not appear in the inner log-likelihood, and we are left with
	\begin{equation}
		\E_{\qr(\vz \mid \vx)} [{\shortminus}\negthinspace\log \pr(\vx \mid \vz)] = \E_{\pr(\vmu_{0})} \E_{\pr(\vmu_{\nrounds} \mid \vmu_{0}, \vx, \bar{\alpha}_{\nrounds})} [{\shortminus}\negthinspace\log \pr(\vx \mid \vz)]. \label{eq:elbo-proof-lr-one}
	\end{equation}
	Since $\lambda_{i} =  \lambda_{0} + \sum_{j=1}^{i} \alpha_{j} = \lambda_{0} + \bar{\alpha}_{i}$, we can define
	\begin{equation}
		\pr(\vmu_{\nrounds} \mid \vmu_{0}, \vx, \lambda) \coloneq \pr(\vmu_{\nrounds} \mid \vmu_{0}, \vx, \alpha = \lambda - \lambda_{0}) = \pr(\vmu_{\nrounds} \mid \vmu_{0}, \vx, \bar{\alpha}_{\nrounds}).
	\end{equation}
	If we now define
	\begin{equation}
		\qr(\vmu_{k} \mid \vx, \lambda_{k}) \coloneq \E_{\pr(\vmu_{0})} \big[\pr(\vmu_{k} \mid \vmu_{0}, \vx, \lambda_{k})\big],
	\end{equation}
	we can rewrite \cref{eq:elbo-proof-lr-one} as
	\begin{equation}
		\E_{\qr(\vz \mid \vx)} [{\shortminus}\negthinspace\log \pr(\vx \mid \vz)] = \E_{\qr(\vmu_{\nrounds} \mid \vx, \lambda_{\nrounds})} [{\shortminus}\negthinspace\log \pr(\vx \mid \vmu_{k})]
	\end{equation}
	which equals the definition of $\lrecon$ after plugging in \cref{eq:elbo-proof-decoder}.

	Next, we investigate the KL-divergence in \cref{eq:elbo-proof-elbo}.
	We begin by defining the latent prior $\pr(\vz)$ autoregressively as
	\begin{equation}
		\pr(\vz) = \pr(\vmu_{0})\,\prod_{i = 1}^{\nrounds} \pr(\vmu_{i} \mid \vmu_{i - 1}, \vxhat_{i - 1}, \alpha_{i}) \label{eq:elbo-proof-latent-prior}
	\end{equation}
	where $\vxhat_{i - 1} = \model(\vmu_{i - 1}, \lambda_{i - 1})$ is the point estimate of $\vx$ produced by our model based on the belief at step $i - 1$.
	So the prior for $\vmu_{i}$ is the update marginal in \cref{thm:update-marginal} if $\vxhat_{i - 1}$ were the actual sample $\vx$.

	Now, we plug \cref{eq:elbo-proof-encoding,eq:elbo-proof-latent-prior} into the KL-divergence term from \cref{eq:elbo-proof-elbo}.
	\begin{equation}
		\begin{aligned}
			\kl(\qr(\vz \mid \vx), \pr(\vz)) & = \E_{\qr(\vz \mid \vx)} \Big[ \log \frac{\qr(\vz \mid \vx)}{\pr(\vz)} \Big]                                                                                                                                                                                                                        \\
			                                 & = \E_{\qr(\vz \mid \vx)} \Big[ \log \frac{\pr(\vmu_{0})}{\pr(\vmu_{0})} + \sum_{i = 1}^{\nrounds} \log \frac{\pr(\vmu_{i} \mid \vmu_{i - 1}, \vx, \alpha_{i})}{\pr(\vmu_{i} \mid \vmu_{i - 1}, \vxhat_{i - 1}, \alpha_{i})} \Big]                                                                   \\
			                                 & = \sum_{i = 1}^{\nrounds} \E_{\qr(\vz \mid \vx)} \Big[ \log \frac{\pr(\vmu_{i} \mid \vmu_{i - 1}, \vx, \alpha_{i})}{\pr(\vmu_{i} \mid \vmu_{i - 1}, \vxhat_{i - 1}, \alpha_{i})} \Big]                                                                                                              \\
			                                 & = \sum_{i = 1}^{\nrounds} \E_{\pr(\vmu_{0})} \E_{\pr(\vmu_{1} \mid \vmu_{0}, \vx, \alpha_{1})}\,\ldots\,\E_{\pr(\vmu_{i} \mid \vmu_{i - 1}, \vx, \alpha_{i})} \Big[ \log \frac{\pr(\vmu_{i} \mid \vmu_{i - 1}, \vx, \alpha_{i})}{\pr(\vmu_{i} \mid \vmu_{i - 1}, \vxhat_{i - 1}, \alpha_{i})} \Big] \\
			                                 & = \sum_{i = 1}^{\nrounds} \E_{\qr(\vmu_{i - 1} \mid \vx, \lambda_{i - 1})} \Big[ \kl\big( \pr(\vmu_{i} \mid \vmu_{i - 1}, \vx, \alpha_{i}), \pr(\vmu_{i} \mid \vmu_{i - 1}, \vxhat_{i - 1}, \alpha_{i}) \big) \Big]
		\end{aligned}
	\end{equation}
	The intermediate expectations have collapsed again according to \cref{thm:multi-update-marginal} in the same way as for the reconstruction term.

	We know the closed form for the inner KL divergences from \cref{thm:kl-div-marginal-update}, so we can further simplify the KL-divergence term to
	\begin{equation}
		\kl(\qr(\vz \mid \vx), \pr(\vz)) = \frac{1}{2} \sum_{i = 1}^{\nrounds} \E_{\qr(\vmu_{i - 1} \mid \vx, \lambda_{i - 1})} \Big[ \alpha_{i} \|\vx - \vxhat_{i - 1}\|_{2}^{2} \Big]. \label{eq:lmeasure-sum-form}
	\end{equation}
	Shifting the sum indices by 1 and replacing the sum $\sum_{i = 0}^{\nrounds - 1}$ with $\nrounds \E_{i \sim \mathcal{U}(0, \nrounds - 1)}$ yields $\lmeasurek$.
\end{proof}

\subsection{\texorpdfstring{Proof of \cref{thm:infinite-elbo}}{Proof of Theorem \ref{thm:infinite-elbo}}}\label{sec:proof-of-infinite-elbo}

\thminfiniteelbo*
\begin{proof}
	Since $\lrecon$ only depends on $\sum_{i} \alpha_{k,i}$ but not individual $\alpha_{k,i}$, the equivalence of the finite- and infinite-step $\lrecon$ is immediately apparent.

	For $\lmeasurek$, we will consider its sum form from \cref{eq:lmeasure-sum-form}.
	\begin{equation}
		\lmeasurek = \frac{1}{2} \sum_{i = 1}^{\nrounds} \E_{\qr(\vmu_{i - 1} \mid \vx, \lambda_{i - 1})} \Big[ \alpha_{i} \|\vx - \vxhat_{i - 1}\|_{2}^{2} \Big] = \frac{1}{2} \sum_{i = 1}^{\nrounds} \alpha_{i} \underbrace{\E_{\qr(\vmu_{i - 1} \mid \vx, \lambda_{i - 1})} \Big[ \|\vx - \vxhat_{i - 1}\|_{2}^{2} \Big]}_{\eqcolon h(\lambda_{i - 1})} \label{eq:inf-elbo-proof-riemann-sum}
	\end{equation}
	Note that $h(\lambda_{i - 1})$ is a deterministic function of $\lambda_{i - 1}$ and $\lambda_{0}, \ldots, \lambda_{\nrounds}$ is a partition of the interval $[\lambda_{0}, \lambda_{0} + \alphaM] = [\lambda_{0}, \lambdaM]$ with interval lengths of $\alpha_{i}$.
	It follows that \cref{eq:inf-elbo-proof-riemann-sum} is a Riemann sum.
	Since $\model$ is a neural network, we can assume that $h$ is continuous almost everywhere.
	Combined with the fact that the interval lengths $\{\alpha_{i}\}$ converge uniformly to $0$, it follows that $\lmeasurek$ converges to the Riemann integral
	\begin{equation}
		\lim_{\nrounds \to \infty} \lmeasurek = \frac{1}{2} \int_{\lambda_{0}}^{\lambdaM} \E_{\qr(\vmu_{\lambda} \mid \vx, \lambda)} \Big[ \|\vx - \vxhat_{\lambda}\|_{2}^{2} \Big]\,\mathrm{d}\lambda \label{eq:elbo-riemann-integral}
	\end{equation}
	as $k \to \infty$.
	It follows trivially that
	\begin{align}
		\lim_{\nrounds \to \infty} \lmeasurek & = \frac{\alphaM}{2} \int_{\lambda_{0}}^{\lambdaM} \frac{1}{\alphaM} \E_{\qr(\vmu_{\lambda} \mid \vx, \lambda)} \Big[ \|\vx - \vxhat_{\lambda}\|_{2}^{2} \Big]\,\mathrm{d}\lambda                                                                                                            \\
		                                      & = \frac{\alphaM}{2} \E_{\substack{\lambda \sim \mathcal{U}(\lambda_{0}, \lambdaM)                                                                                                  \\\qr(\vmu_{\lambda} \mid \vx, \lambda)}} \Big[ \|\vx - \vxhat_{\lambda}\|_{2}^{2} \Big] = \lmeasureinf.
	\end{align}
\end{proof}

\subsection{\texorpdfstring{Proof of \cref{thm:elbo-tighter}}{Proof of Lemma \ref{thm:elbo-tighter}}}\label{sec:proof-of-elbo-tighter}

\thmelbotighter*
\begin{proof}
	In the proof of \cref{thm:infinite-elbo}, we have established that $\lmeasurek$ is a Riemann sum of $h$, where we evaluate $h$ at the left endpoint of each interval.
	Since $h$ is strictly decreasing, $h(\lambda_{i - 1}) > h(\lambda)$ for all $\lambda \in (\lambda_{i - 1}, \lambda_{i}]$ and therefore the discretization error on the interval $[\lambda_{i - 1}, \lambda_{i}]$
	\begin{equation}
		\epsilon \coloneq \alpha_{i}h(\lambda_{i - 1}) - \int_{\lambda_{i - 1}}^{\lambda_{i}} h(\lambda)\,\mathrm{d}\lambda
	\end{equation}
	is strictly positive.
	Summing over all intervals gives the first claim, $\lmeasureinf < \lmeasurek$ for any $k$ and any precision schedule.

	For the second claim, we first consider a refinement of the discretization by a single point $\lambda' \in (\lambda_{i - 1}, \lambda_{i})$ and the post-refinement discretization error on that interval
	\begin{equation}
		\epsilon' \coloneq (\lambda' - \lambda_{i - 1}) h(\lambda_{i - 1}) + (\lambda_{i} - \lambda') h(\lambda') - \int_{\lambda_{i - 1}}^{\lambda_{i}} h(\lambda)\,\mathrm{d}\lambda = (\lambda' - \lambda_{i - 1} - \alpha_{i}) h(\lambda_{i - 1}) + (\lambda_{i} - \lambda') h(\lambda') + \epsilon.
	\end{equation}
	Next, we express $\epsilon'$ in terms of $\epsilon$ as
	\begin{equation}
		\begin{aligned}
			\epsilon' & = (\lambda' - \lambda_{i - 1} - \alpha_{i}) h(\lambda_{i - 1}) + (\lambda_{i} - \lambda') h(\lambda') + \epsilon \\
			          & = (\lambda' - \lambda_{i}) h(\lambda_{i - 1}) + (\lambda_{i} - \lambda') h(\lambda') + \epsilon                  \\
			          & = (\lambda_{i} - \lambda') (h(\lambda') - h(\lambda_{i - 1})) + \epsilon.
		\end{aligned}
	\end{equation}
	We know that $(\lambda_{i} - \lambda') > 0$, because $\lambda' \in (\lambda_{i - 1}, \lambda_{i})$, and $(h(\lambda') - h(\lambda_{i - 1})) < 0$, because $h$ is strictly decreasing.
	It follows that $\epsilon' < \epsilon$.

	The discretization errors on all other intervals are unaffected by this refinement and the integral $\int_{\lambda_{0}}^{\lambdaM} h(\lambda)\,\mathrm{d}\lambda$ does not depend on the schedule at all.
	Therefore, adding a single partition point strictly decreases the total discretization error and thus $\lmeasurek$ itself.
	A refinement in the sense of the lemma adds finitely many partition points, so the second claim follows by induction over these points.
\end{proof}

\subsection{\texorpdfstring{Proof of \cref{thm:elbo-encoder-closed-form} and \cref{thm:bsi-elbo-encoder-closed-form}}{Proof of Lemma \ref{thm:elbo-encoder-closed-form} and Corollary \ref{thm:bsi-elbo-encoder-closed-form}}}\label{sec:proof-prior}

The \elbo{} in \cref{thm:elbo,thm:infinite-elbo} has one part that looks like it might not be so straightforward: the encoding distribution $\qr(\vmu_{\lambda} \mid \vx, \lambda)$.
Its definition contains a marginalization over the belief prior $\pr(\vmu_{0})$, which we still need to specify.
Let's see what $\qr(\vmu_{\lambda} \mid \vx, \lambda)$ becomes if we choose a zero-mean, isotropic Normal prior $\pr(\vmu_{0})$.

\thmelboclosedform*
\begin{proof}
	Let $\pr(\vmu_{\lambda} \mid \vmu_{0}, \vx, \lambda)$ be the marginal distribution of $\vmu_{\lambda}$ after a measurement of precision $\alpha = \lambda - \lambda_{0}$, i.e.
	\begin{equation}
		\pr(\vmu_{\lambda} \mid \vmu_{0}, \vx, \lambda) = \pr(\vmu_{\lambda} \mid \vmu_{0}, \vx, \alpha = \lambda - \lambda_{0}).
	\end{equation}
	We know from \cref{thm:update-marginal} that
	\begin{equation}
		\pr(\vmu_{\lambda} \mid \vmu_{0}, \vx, \alpha = \lambda - \lambda_{0}) = \Np\big( \nicefrac{1}{\lambda} \big[\lambda_{0}\vmu_{0} + (\lambda - \lambda_{0})\vx\big], \nicefrac{\lambda^{2}}{(\lambda - \lambda_{0})} \big).
	\end{equation}
	Since $\pr(\vmu_{0})$ is also Gaussian and $\vmu_{\lambda}$ depends linearly on $\vmu_{0}$, they form a Gaussian linear system for which the marginal distribution of $\vmu_{\lambda}$ is \citep[Equation (4.126)]{murphy2012machine}
	\begin{equation}
		\qr(\vmu_{\lambda} \mid \vx, \lambda) = \E_{\pr(\vmu_{0})} \big[\pr(\vmu_{\lambda} \mid \vmu_{0}, \vx, \lambda)\big] = \N\bigg( \nicefrac{1}{\lambda} \big[\lambda_{0}\vzero + (\lambda - \lambda_0)\vx\big], \frac{\lambda - \lambda_0}{\lambda^{2}} + \frac{\lambda_{0}^{2}}{\lambda^{2}\gamma_{0}} \bigg).
	\end{equation}
	By pulling $\lambda^{-2}$ out of the covariance and inverting to get a precision, we get the claimed result.
\end{proof}

If we now choose $\gamma_{0} = \lambda_{0}$, we get the simple \bsi{} prior and the resulting \elbo{} encoder.

\thmbsielboclosedform*
\begin{proof}
	If we choose $\gamma_{0} = \lambda_{0}$ in \cref{thm:elbo-encoder-closed-form}, we get
	\begin{equation}
		\qr(\vmu_{\lambda} \mid \vx, \lambda) = \Np\bigg( \frac{\lambda - \lambda_{0}}{\lambda}\,\vx, \frac{\lambda^{2}}{\lambda - \lambda_{0} + \nicefrac{\lambda_{0}^{2}}{\lambda_{0}}} \bigg).
	\end{equation}
	The precision simplifies to
	\begin{equation}
		\frac{\lambda^{2}}{\lambda - \lambda_{0} + \nicefrac{\lambda_{0}^{2}}{\lambda_{0}}} = \frac{\lambda^{2}}{\lambda - \lambda_{0} + \lambda_{0}} = \lambda,
	\end{equation}
	proving the result.
\end{proof}

\subsection{\texorpdfstring{Proof of \cref{thm:importance-sampling}}{Proof of Corollary \ref{thm:importance-sampling}}}

\thmimportancesampling*
\begin{proof}
	We know from \cref{eq:elbo-riemann-integral} that $\lmeasureinf$ is the following Riemann integral.
	\begin{equation}
		\lmeasureinf = \frac{1}{2} \int_{\lambda_{0}}^{\lambdaM} \E_{\qr(\vmu_{\lambda} \mid \vx, \lambda)} \Big[ \|\vx - \vxhat_{\lambda}\|_{2}^{2} \Big]\,\mathrm{d}\lambda
	\end{equation}
	Since $\pr$ has support $[\lambda_{0}, \lambdaM]$, we can trivially multiply by $\nicefrac{\pr(\lambda)}{\pr(\lambda)}$ inside the integral and recognize the result as an expectation over $\lambda \sim \pr(\lambda)$, proving the statement.
	\begin{align}
		\lmeasureinf & = \frac{1}{2} \int_{\lambda_{0}}^{\lambdaM} \frac{\pr(\lambda)}{\pr(\lambda)} \E_{\qr(\vmu_{\lambda} \mid \vx, \lambda)} \Big[ \|\vx - \vxhat_{\lambda}\|_{2}^{2} \Big]\,\mathrm{d}\lambda                                                                                                                          \\
		             & = \frac{1}{2} \int_{\lambda_{0}}^{\lambdaM} \pr(\lambda) \E_{\qr(\vmu_{\lambda} \mid \vx, \lambda)} \bigg[ \frac{1}{\pr(\lambda)} \|\vx - \vxhat_{\lambda}\|_{2}^{2} \bigg]\,\mathrm{d}\lambda                                                                                                                      \\
		             & = \frac{1}{2} \E_{\substack{\lambda \sim \pr(\lambda)                                                                                                                                            \\ \qr(\vmu_{\lambda} \mid \vx, \lambda)}} \bigg[ \frac{1}{\pr(\lambda)} \|\vx - \vxhat_{\lambda}\|_{2}^{2} \bigg]
	\end{align}
\end{proof}

\subsection{\texorpdfstring{Proof of \cref{eq:expected-h}}{Proof of Equation \ref{eq:expected-h}}}

\begin{proof}
	We know from \cref{thm:bsi-elbo-encoder-closed-form} that we can write $\vmu_{\lambda} = \nicefrac{\lambda - \lambda_{0}}{\lambda}\,\vx + \nicefrac{1}{\sqrt{\lambda}}\,\vepsilon$ for Gaussian noise $\vepsilon \sim \N(\vzero, \mI)$ independent of $\vx$.
	Together with the assumption $\model(\vmu, \lambda) = \vmu$, we can rewrite $h$ as
	\begin{equation}
		\begin{aligned}
			h(\lambda) & = \E_{\qr(\vmu_{\lambda} \mid \vx, \lambda)} \|\vx - \vxhat_{\lambda}\|_{2}^{2}                                                                                                                                       \\
			           & = \E_{\vepsilon \sim \N(\vzero, \mI)} \bigg\|\vx - \frac{\lambda - \lambda_{0}}{\lambda}\,\vx - \frac{1}{\sqrt{\lambda}}\vepsilon\bigg\|_{2}^{2}                                                                      \\
			           & = \E_{\vepsilon \sim \N(\vzero, \mI)} \bigg\|\frac{\lambda_{0}}{\lambda}\vx - \frac{1}{\sqrt{\lambda}}\vepsilon\bigg\|_{2}^{2}                                                                                        \\
			           & = \E_{\vepsilon \sim \N(\vzero, \mI)} \bigg[{\bigg(\frac{\lambda_{0}}{\lambda}\bigg)}^{2}\|\vx\|_{2}^{2} + \frac{1}{\lambda}\|\vepsilon\|_{2}^{2} - 2 \frac{\lambda_{0}}{\sqrt{\lambda^{3}}}\vx \cdot \vepsilon\bigg].
		\end{aligned}
	\end{equation}
	If we now make use of our assumption that $\E[\vx] = \vzero$ and $\var[\vx] = \mI$, we can distribute the expectation across terms and get
	\begin{equation}
		\E_{\vx}[h(\lambda)] = {\bigg(\frac{\lambda_{0}}{\lambda}\bigg)}^{2} \underbrace{\E_{\vx}\big[\|\vx\|_{2}^{2}\big]}_{= \ndim} + \frac{1}{\lambda}\underbrace{\E_{\vepsilon}\big[\|\vepsilon\|_{2}^{2}\big]}_{= \ndim} - 2 \frac{\lambda_{0}}{\sqrt{\lambda^{3}}}\underbrace{\E_{\vx, \vepsilon}[\vx \cdot \vepsilon]}_{= 0} \propto \frac{\lambda_{0}^{2}}{\lambda^{2}} + \frac{1}{\lambda}.
	\end{equation}
\end{proof}

\section{Experiment Details}\label{sec:experiment-details}

\begin{table}[H]
	\centering
	\caption{Test set log-likelihood on CIFAR10 of the same \unet{} in different models. The \vdm{} and \bfn{} values are the ones reported by \citet{kingma2023variational} and \citet{graves2023bayesian}, respectively, and we train \bsi{} at the corresponding training budget.}\label{tab:cifar10-bpd}
	\begin{tabular}{rcc}
		\toprule
		Model & Training Steps                     & BPD  \\
		\midrule
		VDM   & \multirow{2}{*}{\SI{10}{\million}} & 2.65 \\
		BSI   &                                    & 2.64 \\
		\midrule
		BFN   & \multirow{2}{*}{\SI{5}{\million}}  & 2.66 \\
		BSI   &                                    & 2.65 \\
		\bottomrule
	\end{tabular}
\end{table}

We trained each model on 4 H100 GPUs at a batch size of 128 on CIFAR10 and 512 on ImageNet32 and ImageNet64.
Training progressed at about \num{26300} steps per hour for the \unet{} on CIFAR10 and \num{6100} steps per hour for the \dit{}-L-2 backbones on ImageNet32.
Taking the different batch sizes into account, the two architectures processed about equally many images per hour, \num{3.4} and \num{3.1} million respectively.
Total training time for the \SI{10}{\million} step training on CIFAR10 came to about two weeks.
Since \bfn{}, \vdm{} and \bsi{} share the same backbone and differ only in the encoding distribution, the loss weighting and the sampling update, their per-step training cost and their per-step sampling cost are identical up to measurement noise, so the comparisons in \cref{sec:experiments} are iso-compute.

Furthermore, we take an exponential moving average (EMA) of model weights \citep{song2021scorebased,nichol2021improved}.
We provide an overview of the model and training hyperparameters in \cref{tab:model-params}, and show the \unet{} and \dit{} parameters in \cref{tab:unet-cifar10-params,tab:unet-imagenet32-params,tab:dit-imagenet32-params,tab:dit-imagenet64-params}.
On ImageNet32, we train the models with a cosine learning rate scheduler (with linear warm-up from \num{1e-8}) to achieve faster convergence.
Note that we reduced the training steps to \SI{100}{\thousand} for our parameter studies to make them computationally feasible.

To reduce the variance of the training loss further, we use low-discrepancy sampling for $t$ in \cref{alg:loss} as proposed by \citet{kingma2023variational}.
Instead of sampling $b$ independent $t$ for a batch size of $b$, we set $t_{i} = (\nicefrac{i - 1}{b} + \delta) \bmod 1, i \in [b]$ for a shared $\delta \sim \mathcal{U}(0, 1)$ where $\bmod\ 1$ means that we discard the integer part of the result.
The marginal distribution of each $t_{i}$ is $\mathcal{U}(0, 1)$, but jointly they cover the $[0, 1]$ interval more uniformly than independent samples would, smoothing out the loss across batches.

\begin{table}[H]
	\centering
	\caption{Model and training parameters of BSI on CIFAR10 and all three models on ImageNet32.}\label{tab:model-params}
	\begin{tabular}{rccc}
		\toprule
		                                                       & Parameter               & CIFAR10                             & ImageNet32 (64)                     \\
		\midrule
		\multirow{3}{*}{\rotatebox{90}{BSI}}                   & $\lambda_{0}$           & \multicolumn{2}{c}{\num{1e-2}}                                            \\
		                                                       & $\alphaM$               & \multicolumn{2}{c}{\num{1e6}}                                             \\
		                                                       & $\alphaR$               & \multicolumn{2}{c}{\num{2e6}}                                             \\
		\midrule
		\multirow{5}{*}{\rotatebox{90}{Optim.}}                & Learning rate           & \num{2e-4}                          & \num{5e-4}                          \\
		                                                       & LR Scheduler            & None                                & Cosine $\downarrow$ \num{5e-5}      \\
		                                                       & Weight decay            & \multicolumn{2}{c}{\num{1e-2}}                                            \\
		                                                       & Batch size              & 128                                 & 512                                 \\
		                                                       & Steps                   & \SI{5}{\million}, \SI{10}{\million} & \SI{2}{\million} (\SI{1}{\million}) \\
		\midrule
		\multirow{2}{*}{\smash{\rotatebox[origin=c]{90}{EMA}}} & $\beta$                 & \multicolumn{2}{c}{\num{0.9999}}                                          \\
		                                                       & First update after step & \multicolumn{2}{c}{\num{1000}}                                            \\
		\bottomrule
	\end{tabular}
\end{table}

\begin{minipage}{0.5\textwidth}
	\begin{table}[H]
		\centering
		\caption{\unet{} hyperparameters for CIFAR10.}\label{tab:unet-cifar10-params}
		\begin{tabular}{cc}
			\toprule
			Parameter           & Value \\
			\midrule
			Hidden dim.         & 128   \\
			Levels              & 32    \\
			Dropout             & 0.1   \\
			Attention heads     & 1     \\
			Convolution padding & Zeros \\
			\bottomrule
		\end{tabular}
	\end{table}
\end{minipage}\hfill
\begin{minipage}{0.5\textwidth}
	\begin{table}[H]
		\centering
		\caption{\dit{} hyperparameters for ImageNet32.}\label{tab:dit-imagenet32-params}
		\begin{tabular}{cc}
			\toprule
			Parameter       & Value   \\
			\midrule
			Architecture    & DiT-L-2 \\
			Hidden dim.     & 1024    \\
			Depth           & 24      \\
			Attention heads & 16      \\
			Dropout         & 0.05    \\
			Patch Size      & 2       \\
			\bottomrule
		\end{tabular}
	\end{table}
\end{minipage}

\begin{minipage}{0.5\textwidth}
	\begin{table}[H]
		\centering
		\caption{\unet{} hyperparameters for ImageNet32.}\label{tab:unet-imagenet32-params}
		\begin{tabular}{cc}
			\toprule
			Parameter           & Value \\
			\midrule
			Hidden dim.         & 256   \\
			Levels              & 32    \\
			Dropout             & 0.1   \\
			Attention heads     & 1     \\
			Convolution padding & Zeros \\
			\bottomrule
		\end{tabular}
	\end{table}
\end{minipage}\hfill
\begin{minipage}{0.5\textwidth}
	\begin{table}[H]
		\centering
		\caption{\dit{} hyperparameters for ImageNet64.}\label{tab:dit-imagenet64-params}
		\begin{tabular}{cc}
			\toprule
			Parameter       & Value   \\
			\midrule
			Architecture    & DiT-L-4 \\
			Hidden dim.     & 1024    \\
			Depth           & 24      \\
			Attention heads & 16      \\
			Dropout         & 0.05    \\
			Patch Size      & 4       \\
			\bottomrule
		\end{tabular}
	\end{table}
\end{minipage}

\newpage
\section{Generated Samples}\label{sec:generated-samples}

\begin{figure}[H]
	\centering
	\includegraphics{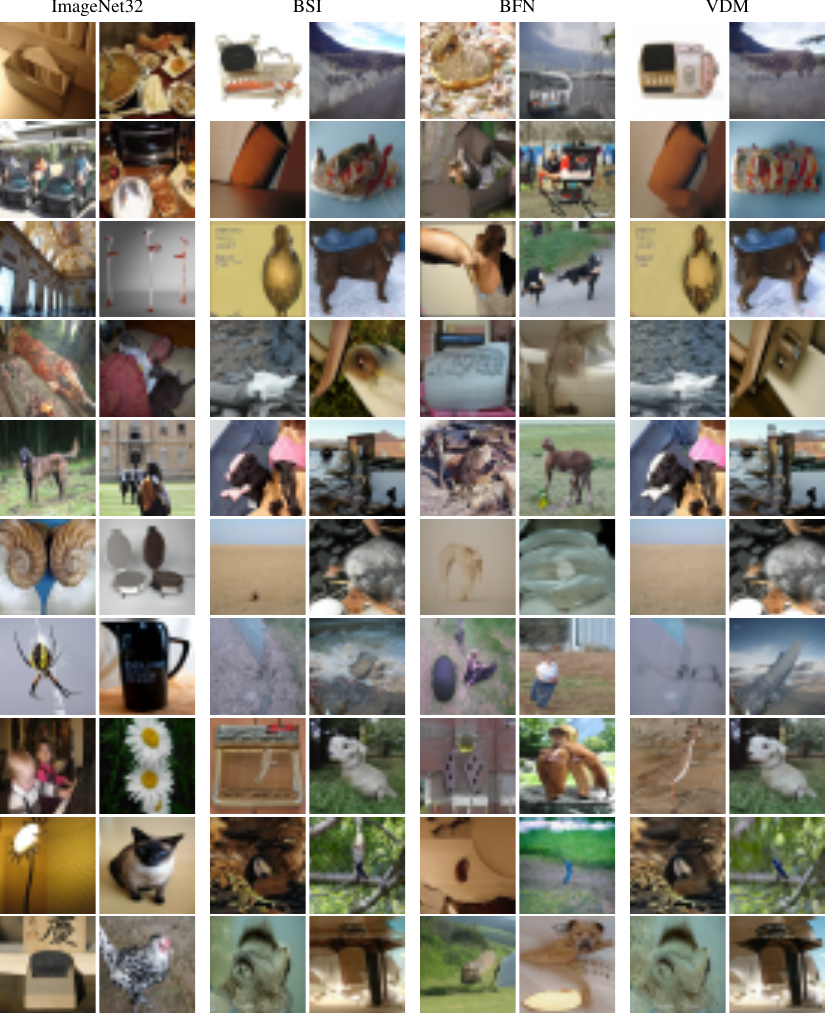}
	\caption{Samples from BSI, BFN and VDM trained on ImageNet32. Generated with 1024 steps. The first two columns show samples from the dataset for comparison.}\label{fig:samples}
\end{figure}

\cref{fig:samples} shows generated samples from models trained on ImageNet32 for visual reference.

\end{document}